\documentclass[journal, 10pt]{IEEEtran}

\usepackage{amsmath,amssymb}
\usepackage{graphicx}
\usepackage{xcolor}
\usepackage{hyperref}
\usepackage{enumitem}
\usepackage{multirow}
\usepackage{algorithm}
\usepackage{algorithmic}
\usepackage{booktabs}
\usepackage{mathtools}

\begin{document}

\title{NaviGNN: Multi-Agent Reinforcement Learning and Graph Neural Network for Sustainable Mobility in Futuristic Smart Cities}

\author{
\begin{center}
{Abderaouf Bahi\textsuperscript{\textbf{1}}} and
{Amel Ourici\textsuperscript{\textbf{2}}}
\\[1.2em]

\textsuperscript{\textbf{1}}\small
Computer Science and Applied Mathematics Laboratory (LIMA)\\ Faculty of Science and Technology, Chadli Bendjedid University, P.O. Box 73, El Tarf 36000, Algeria \\[0.6em]
\textsuperscript{\textbf{2}}\small
Mathematical Modeling and Numerical Simulation Laboratory (LAM2SIN)\\ Faculty of Technology, Badji Mokhtar University, P.O. Box 12, Annaba 23000, Algeria. \\[0.6em]

\textit{*Corresponding author:} Abderaouf Bahi (\textbf{a.bahi@univ-eltarf.dz})
\end{center}
}

\maketitle
\begin{abstract}
This paper investigates the feasibility of human mobility in extreme urban morphologies, characterized by high-density vertical structures and linear city layouts. To assess whether agents can navigate efficiently within such unprecedented topologies, we develop a hybrid simulation framework that integrates agent-based modeling, reinforcement learning (RL), supervised learning, and graph neural networks (GNNs). The simulation captures multi-modal transportation behaviors across multiple vertical levels and varying density scenarios, using both synthetic data and real-world traces from high-density cities. Experiments show that the full AI-integrated architecture enables agents to achieve an average commute time of 7.8--8.4 minutes, a satisfaction rate exceeding 89\%, and a reachability index over 91\%, even during peak congestion periods. Ablation studies indicate that removing intelligent modules such as RL or GNN significantly degrades performance, with commute times increasing by up to 85\% and reachability falling below 70\%. Baseline comparisons against Dijkstra, A*, DQN, and standard GCN further confirm NaviGNN's superiority across all mobility and sustainability metrics. Environmental modeling demonstrates low energy consumption and minimal CO$_2$ emissions when electric modes are prioritized. These results suggest that efficient and sustainable mobility in extreme urban forms is achievable, provided adaptive AI systems, intelligent infrastructure, and real-time feedback mechanisms are implemented.
\end{abstract}

\begin{IEEEkeywords}
Agent-Based Simulation;
Deep Learning;
Human Mobility;
Smart City;
Urban Transportation;
Green Energy
\end{IEEEkeywords}

\small
\begin{table}[htbp!]
\centering
\small
\begin{tabular}{ll}
\hline
\textbf{Symbol} & \textbf{Description} \\
\hline
$N$ & Total number of agents \\
$T_i$ & Travel time of agent $i$ \\
$U_i(m)$ & Utility of agent $i$ for mode $m$ \\
$E_i$ & Energy or effort cost \\
$C_i$ & Context-dependent cost  \\
$T(p), C(p), S(p)$ & Time, congestion cost, preference score of $p$ \\
$w_i$ & Agent-specific weight factors \\
$T_{\text{path}}$ & Travel time of the recommended path \\
$E_{\text{cons}}$ & Energy consumption of the path \\
$U_{\text{agent}}$ & Agent satisfaction score \\
$\alpha,\beta,\gamma$ & Weights for time, energy, and satisfaction \\
$h_v^{(k)}$ & Embedding of node $v$ at layer $k$ \\
$\mathcal{N}(v)$ & Neighbor set of node $v$ \\
$W^{(k)}, b^{(k)}, \sigma$ & GNN weights, bias, and activation \\
$D_{i,j}(t)$ & Predicted density at zone $i$, level $j$, time $t$ \\
$T_t, W_t, E_t$ & Time-of-day, weather, and event indicators \\
$H_{i,j}(t\!-\!1)$ & Historical flow rate \\
$R(s_t,a_t), \gamma$ & Reward function and discount factor \\
$A_{\text{total}}, C_{\text{total}}$ & Total agents and corridor capacity \\
$M, D_j$ & Vertical transfers and delay at transfer $j$ \\
$N_{\text{satisfied}}, N_{\text{total}}$ & Successful and total agents \\
$E, R_k$ & Episodes and reward at episode $k$ \\
$\Delta R_e, \epsilon$ & Reward change and convergence threshold \\
$E', R_{\text{actual}}, R_{\text{optimal}}$ & Perturbed episodes and rewards \\
$C_i^{\text{sel}}, C_i^{\text{sp}}$ & Selected and shortest path costs \\
$l_e, \bar{l}, |E|$ & Edge load, mean load, number of edges \\
$T_{\text{inf}}$ & Inference time per forward pass \\
$P_{\text{mode}}, T_{\text{usage}}$ & Power usage and operation time \\
$EF_{\text{mode}}$ & Emission factor \\
$\mathcal{Z}, T_{i,j}$ & Set of zones and travel time to zone $j$ \\
$\mathbf{A}$ & Graph adjacency matrix \\
$\mathbf{X}$ & Node feature matrix \\
$\hat{\mathbf{A}}$ & Normalized adjacency with self-loops \\
$Q(s,a;\theta)$ & Action-value function parameterized by $\theta$ \\
$\mathcal{B}$ & Replay buffer \\
$\pi_\theta, V_\phi$ & PPO policy and value networks \\
$\hat{A}_t$ & Generalized advantage estimate at step $t$ \\
$\epsilon_{\text{clip}}$ & PPO clipping parameter \\
$\lambda_{\text{GAE}}$ & GAE decay parameter \\
$\rho$ & Agent spatial density (agents/km) \\
$\sigma$ & Road saturation ratio \\
\hline
\end{tabular}
\end{table}

\section{Introduction}

Cities have undergone significant transformations due to population growth, urbanization, and the constant search for smarter, more efficient living environments \cite{1}. In response to these challenges, governments and private actors are envisioning revolutionary urban models that promise to reshape the way we live and move \cite{2}. One of the most ambitious of these projects is NEOM's The Line \cite{3} presented in Figure 1, a proposed linear, car-free smart city in Saudi Arabia stretching over 170 kilometers but only 200 meters wide. This unprecedented design raises a fundamental question: Can we move freely in a linear hyper-dense smart city?

The Line is a core part of Saudi Arabia's Vision 2030 \cite{4}, aiming to accommodate 9 million residents while preserving 95\% of the surrounding natural environment \cite{5}. Unlike traditional cities that expand in concentric or irregular patterns, The Line adopts a linear structure where all services—housing, transportation, and commerce—are vertically integrated within a narrow corridor \cite{6}. The entire city is planned to be powered by renewable energy and connected by high-speed transportation, with no conventional cars or streets. In theory, residents will be able to access their daily needs within a five-minute walk, and travel from one end to the other in 20 minutes via an underground transit system \cite{7}. However, the practicality of human mobility in such a confined and vertically stacked space remains largely unexplored.

This vision brings up an urban paradox. On one hand, The Line promises ecological sustainability, energy efficiency, and technological integration. On the other hand, the hyper-dense, narrow configuration may create new challenges for urban mobility, pedestrian dynamics, and human comfort. Unlike classical cities that allow for radial and grid-based movement, linear cities inherently limit spatial freedom by their geometry. What happens during peak hours when millions of residents try to move across or within this vertical city? Can pedestrian and transport flows coexist without congestion or bottlenecks in such constrained geometry? These open questions are at the heart of our study.

Mobility in urban environments has long been a topic of interest in smart city research \cite{8}. Traditionally, studies have focused on optimizing traffic flows, improving public transport systems, or reducing travel time through urban planning or AI-based approaches \cite{9}. However, these studies mostly concern radial or mesh-like city layouts such as Manhattan or Paris, where multidirectional movement is possible \cite{10}. In contrast, a linear city like The Line forces movement along a single dominant axis. This unique spatial constraint implies a rethinking of traditional mobility models, particularly regarding the feasibility, scalability, and fluidity of movement across different densities and behavioral scenarios.

Moreover, while the official communications emphasize seamless connectivity via ultra-high-speed rail and autonomous transportation, there is limited empirical evidence or simulation-based validation of these claims \cite{11}. Human mobility, especially pedestrian dynamics, cannot be fully predicted without computational modeling. Elements like walking patterns, mode-switching, crowd behavior, queuing effects, and congestion thresholds are critical to assess whether freedom of movement can be maintained in such a dense, constrained environment \cite{12}.

To address this gap, our study proposes a simulation-based evaluation of human mobility in The Line, using agent-based modeling and network simulation to reproduce and analyze movement dynamics within a hypothetical segment of the city. Our simulation incorporates multiple layers of mobility: pedestrian flow, transit systems, elevators/escalators, and congestion factors. We simulate different population densities, time periods (rush hour vs. off-peak), and agent behavior rules to measure key performance indicators such as travel time, path blockage, and density-pressure relationships. These simulations aim to assess both micro (individual) and macro (city-wide) mobility patterns and determine whether the promise of seamless movement holds under realistic conditions.

Our proposed simulation framework builds on concepts from urban mobility modeling, complex networks, and smart city planning \cite{13}. It integrates behavioral variability among agents, heterogeneous destinations, and the architectural constraints of a linear smart city. This allows us to provide a quantifiable and visual answer to the core research question: Can we move freely in The Line?

To summarize, this work offers the following contributions:

\begin{itemize}
    \item A novel simulation of human mobility in NEOM's The Line, representing the first systematic attempt to evaluate mobility feasibility in a hyper-dense linear smart city.
    \item A formal mathematical framework (Section~3) grounding the agent decision model, graph topology, and RL formulation in rigorous preliminaries.
    \item An agent-based modeling framework incorporating multiple layers of mobility (walking, vertical transport, high-speed transit) and variable density scenarios.
    \item A quantitative evaluation of movement freedom, based on indicators such as average travel time, flow saturation, and space utilization efficiency.
    \item A comparative baseline study benchmarking NaviGNN against classical and learning-based routing approaches (Dijkstra, A*, DQN, standard GCN).
    \item Insightful discussion on the implications of linear geometry on human behavior, urban planning, and smart infrastructure design, with a focus on scalability and livability.
\end{itemize}

The rest of this paper is structured as follows: Section 2 presents the related work on urban mobility modeling and smart city simulations. Section 3 introduces the mathematical preliminaries underlying our framework. Section 4 details the methodology. Section 5 presents the results and analysis of different scenarios. Finally, Section 6 concludes the paper with key takeaways and recommendations.

\begin{figure*}[htbp]
\centering
\includegraphics[width=1\linewidth]{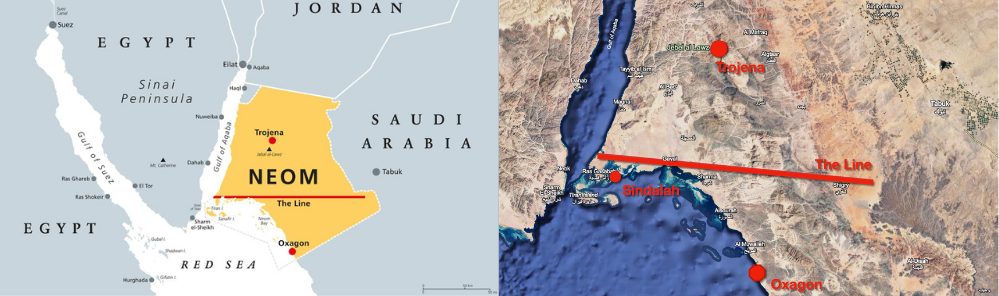}
\caption{NEOM's The Line Chart}
\end{figure*}

\section{Related works}

The growing interest in sustainable urban development has led to numerous studies exploring smart mobility systems within emerging city models. Smart cities aim to integrate technology, data, and sustainable infrastructure to enhance the efficiency and quality of urban living. In this context, \cite{14} highlights the critical role of digital twins and spatial computing in reimagining urban spaces, laying the conceptual foundation for futuristic cities like NEOM. Similarly, \cite{15} provides an overview of artificial intelligence applications across NEOM's sectors, emphasizing the integration of smart mobility as a foundational service. These visionary perspectives are supported by \cite{16}, which stresses the importance of high-resolution simulation models in smart cities to capture complex urban dynamics and inform decision-making. Furthermore, \cite{17} discusses digital twin-enabled urban computing as a powerful framework to optimize city planning, especially concerning traffic and public transport.

Despite their visionary scope, these contributions remain largely conceptual and lack granular modeling or simulation of individual mobility behaviors. They typically focus on high-level infrastructures or technologies rather than the micro-level modeling required to simulate population flows in an elongated, linear city.

To simulate how individuals move within cities, researchers have explored both Agent-Based Models (ABM) and machine learning frameworks. ABM allows for modeling mobility patterns by simulating the actions and interactions of autonomous agents. \cite{18} proposes an ABM that captures the emergence of urban mobility based on real-time data, demonstrating the potential of such methods to reflect complex behaviors in a dynamic environment. Complementing this, \cite{19} employs a multi-agent system to simulate urban mobility scenarios in Brazil, integrating geographic and transport data layers for realism. These studies suggest that ABMs are well-suited to model large-scale individual movements, especially in newly planned environments with sparse historical data.

Meanwhile, machine learning approaches have gained popularity for their predictive accuracy. \cite{20} compares ML and deep learning \cite{21,22} methods to model urban mobility patterns and forecasts, concluding that while ML models perform well, they often require large historical datasets—something that might be lacking in brand-new cities like NEOM. \cite{23} offers a hybrid model that integrates public transport data with individual movement patterns using deep learning, showing improved accuracy in short-term forecasting. However, both studies emphasize the need for substantial, historical data, which may limit their applicability to futuristic city projects.

Although both ABM and ML offer strengths, neither approach fully addresses the simulation needs of a linear city with novel constraints like NEOM. ABM offers adaptability, while ML requires training data that may not exist in such a novel setting.

A key component of mobility modeling lies in understanding how people interact with buildings—where they live, work, and engage in daily activities. \cite{24} introduces a building-centric model that generates synthetic populations and assigns them activities based on land use and floor-area ratios. This approach provides a foundation for simulating urban dynamics in the absence of real demographic data—an issue that NEOM will face during its early development stages. However, the model assumes traditional grid-based city layouts and does not consider the unique constraints of a 170 km linear structure, such as spatial bottlenecks or highly constrained path choices.

While many studies envision NEOM as a testbed for futuristic urban development, few have attempted to simulate its mobility dynamics in detail. \cite{25} introduces a generative design platform for NEOM's transportation planning, using optimization to balance land use and accessibility. However, it does not model individual mobility behaviors or micro-interactions. \cite{26} reviews autonomous vehicle integration within The Line, proposing that shared AV fleets could serve as the main transport mode, but this remains hypothetical and lacks behavioral validation. \cite{27} provides a strategic overview of sustainable infrastructure in NEOM, focusing on water, energy, and transport integration. While insightful, it again lacks simulation components. Lastly, \cite{28} explores NEOM's zero-carbon goals through transport planning but provides no behavioral modeling of users navigating its unique linear shape.

These works reveal a clear gap: while the vision and infrastructure of The Line are well-documented, there remains a lack of models that simulate how residents would move, choose destinations, and experience space on a day-to-day basis. 

The existing literature presented in Table~\ref{tab:soa} showcases a rich body of work on smart cities, mobility modeling, and conceptual planning for NEOM. However, none of the current studies fully address the simulation of human mobility within a linear city constrained by a single spine of transit, as is the case in The Line. Our work addresses this gap by proposing a building-informed agent-based model tailored to the linear structure of The Line, where agents' movement is shaped by building types, temporal activities, and proximity to the high-speed transit.

\begin{table*}[htbp!]
\centering
\caption{State of art summary}
\label{tab:soa}
\small
\begin{tabular}
{p{0.5cm}p{0.5cm}p{2.5cm}p{3.5cm}p{3cm}p{2cm}}
\hline
Ref & Year & Approach & Contribution & Limitation & Dataset  \\
\hline
{\cite{14}} & 2023 & Conceptual (Digital Twin) & Spatial computing for futuristic cities & No simulation or mobility modeling & Conceptual \\
{\cite{15}} & 2022 & Review & AI integration in NEOM & High-level only & NEOM overview \\
{\cite{16}} & 2021 & Simulation Framework & Smart city digital simulation & No ABM or individual models & Smart cities \\
{\cite{17}} & 2021 & Urban Computing & Digital twins for urban optimization & No behavioral mobility modeling & General \\
{\cite{18}} & 2020 & ABM & Real-time mobility emergence & Needs real-time data & Urban environment \\
{\cite{19}} & 2021 & ABM & Multi-agent system for Brazilian city & Location-specific & Brazilian dataset \\
{\cite{20}} & 2022 & ML/DL & Comparison of ML and DL for mobility & Requires large datasets & Generic urban data \\
{\cite{21}} & 2023 & Hybrid DL + Public Transport & Mobility prediction & Needs historical transit data & Transit cities \\
{\cite{22}} & 2020 & Building-Centric Simulation & Synthetic population \& activity assignment & Grid-based only & U.S. urban layouts \\
{\cite{23}} & 2022 & Optimization & Transport-access design in NEOM & No micro-simulation & NEOM transport \\
{\cite{24}} & 2022 & Conceptual (AV Systems) & Autonomous mobility for The Line & No user modeling & AV scenarios \\
{\cite{25}} & 2021 & Strategic Review & Sustainable infrastructure planning & No individual mobility focus & NEOM strategy \\
{\cite{26}} & 2023 & Transport Strategy & Zero-carbon vision for NEOM & No simulation or agents & Vision-based \\
\hline
\end{tabular}
\end{table*}

\section{Preliminaries}

This section establishes the formal mathematical foundations that underpin the NaviGNN framework. We define the graph representation of The Line, formalize the Markov Decision Process (MDP) governing agent navigation, and introduce the key properties of the Graph Neural Network used for route optimization.

\subsection{Graph Representation of the Urban Network}

We model The Line as a directed, weighted, multilayer graph
\begin{equation}
    \mathcal{G} = (\mathcal{V},\, \mathcal{E},\, \mathbf{X},\, \mathbf{W}),
\end{equation}
where $\mathcal{V} = \{v_1, v_2, \ldots, v_{|\mathcal{V}|}\}$ is the set of nodes (urban zones or transit stops), $\mathcal{E} \subseteq \mathcal{V} \times \mathcal{V}$ is the set of directed edges (corridors, lifts, or transit links), $\mathbf{X} \in \mathbb{R}^{|\mathcal{V}| \times F}$ is the node feature matrix with $F$ features per node, and $\mathbf{W} \in \mathbb{R}^{|\mathcal{E}|}$ is the edge-weight vector encoding travel cost.

The adjacency matrix $\mathbf{A} \in \{0,1\}^{|\mathcal{V}| \times |\mathcal{V}|}$ has entry $A_{uv} = 1$ if a directed link $(u \to v) \in \mathcal{E}$ exists, and $0$ otherwise. The degree matrix is $\mathbf{D} = \operatorname{diag}(d_1, \ldots, d_{|\mathcal{V}|})$ where $d_v = \sum_u A_{vu}$.

The normalized adjacency with added self-loops is
\begin{equation}
    \hat{\mathbf{A}} = \tilde{\mathbf{D}}^{-1/2}\,\tilde{\mathbf{A}}\,\tilde{\mathbf{D}}^{-1/2},
    \quad \tilde{\mathbf{A}} = \mathbf{A} + \mathbf{I},
    \quad \tilde{D}_{vv} = \sum_u \tilde{A}_{vu}.
\end{equation}

Each node $v \in \mathcal{V}$ carries a feature vector
\begin{equation}
    \mathbf{x}_v = [z_v,\; \ell_v,\; \rho_v(t),\; \kappa_v,\; \eta_v]^\top \in \mathbb{R}^F,
\end{equation}
where $z_v$ is the zone type (residential, commercial, transit, recreational), $\ell_v$ is the vertical level index, $\rho_v(t)$ is the current occupancy density at time $t$, $\kappa_v$ is the corridor capacity, and $\eta_v \in [0,1]$ is the current saturation ratio. The edge weight between nodes $u$ and $v$ is a composite cost:
\begin{equation}
    w_{uv}(t) = \alpha_1\,\tau_{uv} + \alpha_2\,\rho_v(t)\,\kappa_v^{-1} + \alpha_3\,e_{uv},
\end{equation}
where $\tau_{uv}$ is the nominal travel time, $\rho_v(t)\,\kappa_v^{-1}$ is the congestion penalty, and $e_{uv}$ is the energy cost of traversing the link, with $\alpha_1 + \alpha_2 + \alpha_3 = 1$.

The multilayer structure of The Line is captured by partitioning $\mathcal{V}$ into $L = 50$ level-specific subsets $\mathcal{V}_\ell$ and augmenting $\mathcal{E}$ with vertical transition edges $\mathcal{E}^{\uparrow\downarrow}$ modeling elevator and escalator links:
\begin{equation}
    \mathcal{E} = \mathcal{E}^{\text{horiz}} \cup \mathcal{E}^{\uparrow\downarrow},
    \quad \mathcal{E}^{\uparrow\downarrow} = \{(v_\ell, v_{\ell+1}) \mid v \in \mathcal{V},\; 1 \le \ell < L\}.
\end{equation}

\subsection{Markov Decision Process Formulation}

Agent navigation is formalized as a discrete-time Markov Decision Process (MDP) defined by the tuple $\mathcal{M} = (\mathcal{S}, \mathcal{A}, \mathcal{P}, \mathcal{R}, \gamma)$.

\paragraph{State Space} The state of agent $i$ at time step $t$ is
\begin{equation}
    s_t^i = \bigl(v_t^i,\; d_t^i,\; \rho_{v_t^i}(t),\; \sigma_{v_t^i}(t),\; m_t^i,\; \tau_{\text{rem}}^i \bigr),
\end{equation}
where $v_t^i \in \mathcal{V}$ is the current node, $d_t^i \in \mathcal{V}$ is the destination node, $\rho_{v_t^i}(t)$ is the local density, $\sigma_{v_t^i}(t)$ is the saturation ratio, $m_t^i$ is the current transport mode, and $\tau_{\text{rem}}^i$ is the remaining time budget.

\paragraph{Action Space} The action set of agent $i$ at node $v$ is
\begin{equation}
    \mathcal{A}(v) = \bigl\{(u, m) \mid (v, u) \in \mathcal{E},\; m \in \mathcal{M}_v \bigr\},
\end{equation}
where $\mathcal{M}_v$ is the set of transport modes available at node $v$ (pedestrian, cyclist, shuttle, drone).

\paragraph{Transition Dynamics} The stochastic transition kernel is
\begin{equation}
    \mathcal{P}(s_{t+1}^i \mid s_t^i, a_t^i) = \Pr\!\bigl[s_{t+1}^i = s' \mid s_t^i = s,\; a_t^i = a\bigr],
\end{equation}
which depends on congestion dynamics and the probabilistic delays induced by corridor saturation.

\paragraph{Reward Function} The reward issued after action $a_t^i = (u, m)$ is
\begin{equation}
    R(s_t^i, a_t^i) = -\lambda_1\,\Delta\tau_t^i - \lambda_2\,\sigma_{u}(t) + \lambda_3\,\mathbf{1}[v_t^i = d_t^i] - \lambda_4\,e_{v_t^i u},
\end{equation}
where $\Delta\tau_t^i = \tau_{v_t^i u}(t)$ is the incurred travel time, $\sigma_u(t)$ penalizes congested destinations, $\mathbf{1}[v_t^i = d_t^i]$ provides a terminal bonus, and $e_{v_t^i u}$ penalizes energy-intensive actions. The weights $\lambda_k > 0$ satisfy $\sum_k \lambda_k = 1$.

\paragraph{Return and Optimal Policy} The expected discounted return from step $t$ is
\begin{equation}
    G_t^i = \mathbb{E}\!\left[\sum_{k=0}^{T-t} \gamma^k R(s_{t+k}^i, a_{t+k}^i)\right], \quad \gamma \in [0,1).
\end{equation}
The optimal policy is $\pi^*(a \mid s) = \arg\max_\pi \mathbb{E}_\pi[G_0^i]$.

\subsection{Deep Q-Network (DQN) Formulation}

The action-value function $Q^\pi(s, a) = \mathbb{E}_\pi[G_t \mid s_t = s, a_t = a]$ satisfies the Bellman optimality equation:
\begin{equation}
    Q^*(s,a) = \mathbb{E}\!\left[R(s,a) + \gamma \max_{a'} Q^*(s', a')\right].
\end{equation}
We approximate $Q^*$ with a deep neural network $Q(s, a;\theta)$, trained by minimizing the temporal-difference (TD) loss:
\begin{equation}
    \mathcal{L}_{\text{DQN}}(\theta) = \mathbb{E}_{(s,a,r,s') \sim \mathcal{B}}\!\left[\bigl(y - Q(s,a;\theta)\bigr)^2\right],
\end{equation}
where $y = r + \gamma \max_{a'} Q(s', a';\theta^-)$ is the target value and $\theta^-$ are the parameters of a periodically synchronized target network. Experiences $(s, a, r, s')$ are stored in a replay buffer $\mathcal{B}$ and sampled uniformly to break temporal correlations.

\subsection{Proximal Policy Optimization (PPO)}

For cooperative agents (e.g., autonomous shuttle fleets), we employ PPO, which optimizes a clipped surrogate objective to ensure stable policy updates:
\begin{equation}
    \mathcal{L}_{\text{PPO}}(\theta) = \mathbb{E}_t\!\left[\min\!\left(r_t(\theta)\hat{A}_t,\; \operatorname{clip}\!\left(r_t(\theta), 1\!-\!\epsilon_{\text{clip}}, 1\!+\!\epsilon_{\text{clip}}\right)\hat{A}_t\right)\right],
\end{equation}
where $r_t(\theta) = \pi_\theta(a_t \mid s_t) / \pi_{\theta_{\text{old}}}(a_t \mid s_t)$ is the probability ratio and $\hat{A}_t$ is the Generalized Advantage Estimate (GAE):
\begin{equation}
    \hat{A}_t = \sum_{l=0}^{T-t-1} (\gamma\lambda_{\text{GAE}})^l \,\delta_{t+l},
    \quad \delta_t = r_t + \gamma V_\phi(s_{t+1}) - V_\phi(s_t).
\end{equation}
The value network $V_\phi$ is updated by minimizing:
\begin{equation}
    \mathcal{L}_{\text{val}}(\phi) = \mathbb{E}_t\!\left[\bigl(V_\phi(s_t) - \hat{G}_t\bigr)^2\right],
\end{equation}
where $\hat{G}_t$ is the empirical return. The joint PPO loss is:
\begin{equation}
    \mathcal{L}(\theta,\phi) = -\mathcal{L}_{\text{PPO}}(\theta) + c_1\,\mathcal{L}_{\text{val}}(\phi) - c_2\,\mathcal{H}[\pi_\theta],
\end{equation}
with entropy bonus $\mathcal{H}[\pi_\theta]$ encouraging exploration.

\subsection{Graph Neural Network for Routing}

The GNN processes the graph $\mathcal{G}$ to produce congestion-aware node embeddings. At layer $k$, node $v$ aggregates messages from its neighborhood:
\begin{equation}
    \mathbf{m}_v^{(k)} = \operatorname{AGG}^{(k)}\!\left(\left\{W_e^{(k)}\,h_u^{(k-1)} \mid u \in \mathcal{N}(v),\; e=(u,v)\right\}\right),
\end{equation}
where $W_e^{(k)}$ is an edge-conditioned weight matrix. The node representation is updated as:
\begin{equation}
    h_v^{(k)} = \sigma\!\left(W^{(k)}\!\left[h_v^{(k-1)} \,\|\, \mathbf{m}_v^{(k)}\right] + b^{(k)}\right),
\end{equation}
with $\|$ denoting concatenation. The initial embeddings are $h_v^{(0)} = \mathbf{x}_v$.

For route scoring, a path $p = (v_0, v_1, \ldots, v_K)$ receives a composite score:
\begin{equation}
    \text{score}(p) = \sum_{k=1}^{K} w_{v_{k-1} v_k}\!\left(h_{v_{k-1}}^{(L)}, h_{v_k}^{(L)}\right),
\end{equation}
and the optimal path is selected as $p^* = \arg\min_p \text{score}(p)$. The GNN training objective combines the routing loss with a regularization term:
\begin{equation}
    \mathcal{L}_{\text{GNN}} = \underbrace{T_{\text{path}} + \beta E_{\text{cons}} + (1-U_{\text{agent}})}_{\text{routing loss}} + \mu\,\|\Theta_{\text{GNN}}\|_F^2,
\end{equation}
where $\Theta_{\text{GNN}}$ collects all GNN parameters and $\|\cdot\|_F$ is the Frobenius norm.

\subsection{Demand Forecasting via Supervised Learning}

The spatial-temporal density prediction is modeled as a multivariate regression:
\begin{equation}
    \hat{D}_{i,j}(t) = f_\theta\!\left(\mathbf{c}_{i,j}(t)\right),
    \quad \mathbf{c}_{i,j}(t) = \bigl[T_t,\; W_t,\; E_t,\; H_{i,j}(t{-}\tau),\; \rho_{i,j}(t{-}1)\bigr],
\end{equation}
trained with the Huber loss to handle outliers:
\begin{equation}
    \mathcal{L}_{\text{Huber}}(\hat{D}, D) = \begin{cases}
    \tfrac{1}{2}(\hat{D} - D)^2 & |\hat{D} - D| \le \delta, \\
    \delta\!\left(|\hat{D} - D| - \tfrac{\delta}{2}\right) & \text{otherwise.}
    \end{cases}
\end{equation}

\section{Methodology}

The proposed simulation and decision-making system shown in Figure~2 is structured around a modular architecture that integrates three core components: the Agent-Based Core Simulator (ABCS), the AI Decision Layer (AIDL), and the Mobility Graph Model (MGM). Each module plays a distinct role in modeling, controlling, and optimizing mobility behaviors within the simulated environment of The Line.

The Agent-Based Core Simulator (ABCS) is implemented using a hybrid of Mesa (for high-level agent logic and interaction modeling) and SUMO (Simulation of Urban Mobility) for detailed traffic flow and transportation dynamics. It manages the spatial-temporal evolution of thousands of agents—pedestrians, cyclists, shuttles, and drones—across horizontal and vertical layers of the city, simulating real-world constraints such as elevator delays, corridor saturation, and travel schedules.

On top of this simulation backbone, the AI Decision Layer (AIDL) governs agent-level decision-making. It integrates Reinforcement Learning (RL) policies—trained using Deep Q-Learning and Proximal Policy Optimization (PPO)—to allow autonomous shuttles and other agents to adaptively select optimal routes based on evolving traffic and environmental conditions. Additionally, it includes a Supervised Learning module, powered by XGBoost, to forecast commuter demand and identify service bottlenecks. These predictions are injected periodically into the agent environment to influence behavior, e.g., rerouting traffic before a peak or dispatching additional autonomous vehicles.

Complementing these systems is the Mobility Graph Model (MGM), which uses a Graph Neural Network (GNN) to represent and process the city's complex network topology. The GNN computes a latent embedding for each node (zone, corridor, or level) and learns optimal routing strategies based on load balancing, path efficiency, and energy consumption.

A custom middleware pipeline ensures real-time interaction between components. Intermediate outputs from SUMO are fed into the GNN to update traffic and network representations. The GNN, in turn, passes optimized routing recommendations to the RL agents, enabling adaptive planning under dynamic and stochastic conditions. This pipeline ensures a high degree of fidelity and reactivity, allowing the system to simulate emergent behaviors with fine-grained accuracy across temporal phases and agent categories.

\subsection{Hypothetical Model of The Line}

This section introduces a conceptual model of The Line, a futuristic linear city proposed in Saudi Arabia, which serves as the foundational framework for our agent-based simulation study. The model envisions The Line as a 170-kilometer-long, 200-meter-wide, and 500-meter-tall urban corridor. Its vertical dimension allows for approximately 50 levels, effectively stacking multiple urban functionalities to minimize land usage while maximizing efficiency.

The internal layout of the city is designed with a functional distribution that reflects a balanced, sustainable urban ecology: approximately 40\% of the space is allocated to residential zones, 30\% to commercial and service areas, 15\% to transportation corridors, and the remaining 15\% to public spaces and green infrastructure. These zones are not spread horizontally as in traditional cities, but are integrated vertically, with each layer optimized for a specific function.

The city is further stratified into mobility and service layers, allowing for dedicated vertical zoning to separate pedestrian activity, vehicular transport, and service delivery. This vertical partitioning aims to reduce surface congestion, improve accessibility, and facilitate real-time routing of autonomous systems. Table~2 illustrates the vertical functional distribution across the 50 levels, which guides the spatial behavior of agents in our simulation. The 2D schematic in Figure~3 visualizes pedestrian agents, cyclists, autonomous shuttles, drones, and vertical connectors as they navigate the multi-layered structure of The Line.

\begin{table}[htbp]
\centering
\caption{Vertical functional allocation of The Line}
\small
\begin{tabular}{ll}
\hline
Level & Functionality \\
\hline
1--10 & Pedestrian and light vehicle movement \\
11--20 & Residential zones \\
21--30 & Commercial and educational hubs \\
31--40 & Healthcare and recreational areas \\
41--50 & Autonomous high-speed transit systems \\
\hline
\end{tabular}
\end{table}

This architecture facilitates reduced travel time, minimized carbon footprint, and maximized space usage, forming the backbone of our simulation scenario.

\begin{figure}[htbp]
\centering
\includegraphics[width=1\linewidth]{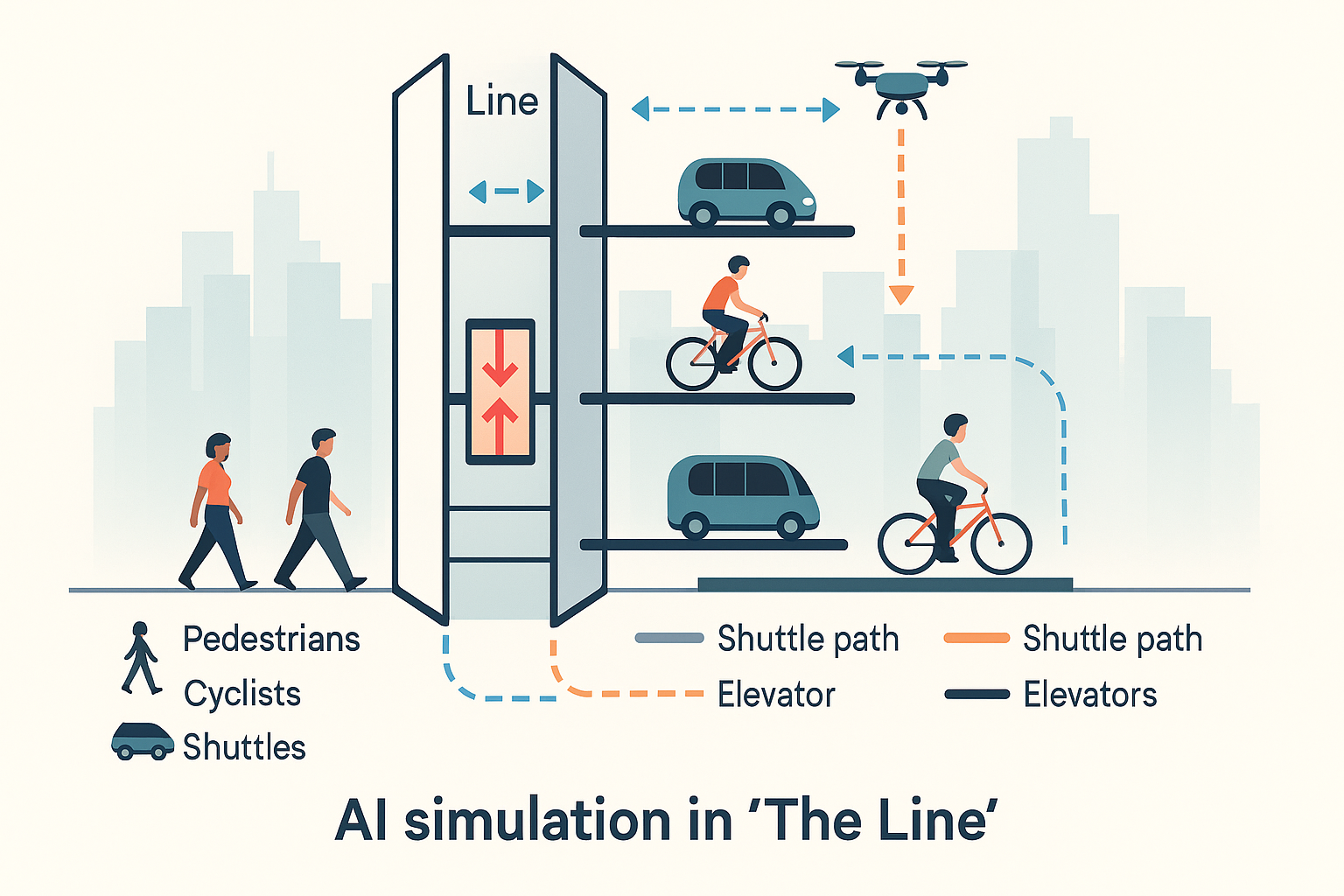}
\caption{Agent-based simulation of human mobility in The Line}
\end{figure}

Although the official NEOM plans emphasize a high-speed underground spine and walkable urban layers, our simulation introduces autonomous shuttles to explore scenarios that address last-mile mobility, accessibility, and potential transitional needs during The Line's early operation phases.

\subsection{Simulation Scenario}

To model the transportation and mobility dynamics within The Line, we construct a detailed simulation scenario that incorporates both spatial and temporal elements, aimed at capturing the unique features of this vertically integrated smart city. The urban structure is abstracted into a 2D grid-based layout, where each grid cell represents a horizontal section of the city. Vertical complexity is addressed through hierarchical layers, simulating the 50 functional levels discussed previously in Table~2. This hybrid 2D-vertical representation allows us to manage both lateral and vertical mobility, including elevator use, multi-level transitions, and intermodal interactions.

Four primary agent types are defined in the simulation to reflect realistic urban mobility patterns: pedestrians (including walkers and joggers), cyclists (both electric and manual), autonomous shuttles (used for intra-city public and private transport), and delivery drones (which model logistics and last-mile delivery behavior). Each agent type has distinct speed, energy consumption, and behavioral characteristics, enabling diverse interaction profiles and congestion patterns.

To reflect the temporal variability of urban mobility, we simulate three distinct time periods across a standard day: morning rush hour (7 AM to 9 AM), midday low-traffic period (12 PM to 2 PM), and evening rush hour (5 PM to 7 PM). These periods are chosen to study peak and off-peak behavior, allowing us to measure stress points in mobility infrastructure such as elevator delays and corridor saturation. The average travel distance within The Line is estimated to range between 2 to 5 kilometers per trip, varying by agent type and individual objective (e.g., commuting, recreation, delivery).

The intensity and fluctuation of agent flows across these periods are summarized in Table~3, which captures the peak and off-peak traffic loads for each mobility category. This information is used to initialize the simulation environment and serves as a benchmark for validating the emergence of realistic traffic and congestion dynamics.

\begin{table}[htbp]
\centering
\caption{Expected agent flow rates by type and time period}
\small
\begin{tabular}{lcc}
\hline
Agent Type & Peak Flow (agents/hr) & Off-Peak Flow (agents/hr) \\
\hline
Pedestrians & 10,000 & 4,000 \\
Cyclists & 2,500 & 800 \\
Shuttles & 1,200 & 500 \\
Drones & 600 & 300 \\
\hline
\end{tabular}
\end{table}

This simulation scenario is critical for exploring the complex interplays between vertical mobility, agent density, and infrastructure constraints, offering insights into how The Line may perform under realistic urban demand conditions.

To accurately capture the complex micro- and macro-level mobility patterns inherent in The Line, we implement a robust agent-based simulation framework tailored to this linear, multi-level urban environment. The simulation relies on a combination of computational tools to reflect the heterogeneity of agents, spatial layering, and multimodal transportation systems. We utilize Mesa, a Python-based modular framework well-suited for developing and managing agent-based models at scale, enabling fine-grained customization of agent behaviors and interactions. For traffic-specific modeling, SUMO (Simulation of Urban Mobility) is integrated to simulate vehicle flows and autonomous shuttle operations, particularly along designated transport corridors. NetLogo is employed in early stages for prototyping and interactive validation of core mobility rules, ensuring model correctness before deployment at scale. Additionally, NetworkX, a Python library for graph-based modeling, is used to structure The Line as a multi-layered network, representing zones and interconnections across levels through directed weighted graphs.

Each simulation run is initialized with a carefully defined set of parameters. These parameters are summarized in Table~4. For instance, agent density ($\rho$) ranges from 10 to 500 agents per kilometer, depending on the simulation scenario and time of day. Walking speed ($v_w$) and cycling speed ($v_c$) are set within empirically observed ranges for urban settings, allowing us to simulate various human mobility patterns. Shuttle speeds ($v_s$) are parameterized between 5 and 15 m/s, reflecting low- to mid-speed electric transport vehicles typically proposed for urban use. A unique constraint of The Line—vertical mobility—is incorporated via the vertical lift time ($t_l$) parameter, which varies between 10 to 60 seconds depending on elevator efficiency and agent traffic. Lastly, road saturation ($\sigma$) is modeled as a normalized variable from 0 to 1, representing the extent of space utilization in transport corridors, which becomes critical for understanding congestion thresholds.

\begin{table*}[htbp]
\centering
\caption{Simulation parameterization}
\small
\begin{tabular}{lll}
\hline
Parameter & Value Range & Description \\
\hline
Agent Density ($\rho$) & 10 to 500 agents/km & Number of agents per kilometer segment \\
Walking Speed ($v_w$) & 0.8 to 1.5 m/s & Pedestrian walking speed \\
Cycling Speed ($v_c$) & 3 to 7 m/s & Cyclist average speed \\
Shuttle Speed ($v_s$) & 5 to 15 m/s & Autonomous shuttle cruising speed \\
Vertical Lift Time ($t_l$) & 10 to 60 sec & Time to switch between levels via elevators \\
Road Saturation ($\sigma$) & 0 to 1 (normalized) & Ratio of used capacity to total corridor capacity \\
\hline
\end{tabular}
\end{table*}

\subsection{Implementation}

To simulate and optimize transportation and mobility within The Line, we leverage Artificial Intelligence techniques that enable adaptive, dynamic, and realistic behaviors of agents. These techniques enhance the fidelity and decision-making capacity of agents in the simulation. The following methods are used:

\subsubsection{Reinforcement Learning (RL) for Autonomous Navigation}

Reinforcement Learning (RL) plays a central role in enabling adaptive and intelligent navigation strategies for autonomous mobility agents within The Line, such as self-driving shuttles and aerial delivery drones. Specifically, advanced RL algorithms like Deep Q-Learning (DQL) and Proximal Policy Optimization (PPO) are employed to help these agents learn optimal navigation policies in complex, dynamic environments. The agents continuously interact with the simulated cityscape, receiving feedback in the form of rewards or penalties based on the efficiency, safety, and timeliness of their travel routes.

The learning process is governed by a reward structure that encapsulates critical urban mobility objectives such as minimizing travel time, avoiding congestion, and adhering to energy efficiency constraints. Over time, the RL agent converges to a policy $\pi^*$ that maximizes the expected cumulative reward across different mobility contexts. The expected reward function is expressed in Equation~(1).

\begin{equation}
\pi^* = \arg\max \mathbb{E} \left[ \sum_{t=0}^{T} \gamma^t R(s_t, a_t) \right]
\end{equation}

Algorithm~\ref{alg:navignn} presents the full NaviGNN training loop, integrating the GNN routing module, the RL policy update, and the ABCS feedback cycle.

\begin{algorithm}[htbp]
\caption{NaviGNN: Integrated Training Loop}
\label{alg:navignn}
\begin{algorithmic}[1]
\REQUIRE Graph $\mathcal{G}=(\mathcal{V},\mathcal{E},\mathbf{X},\mathbf{W})$, agent set $\mathcal{I}$, episodes $E$, steps per episode $T$
\REQUIRE GNN parameters $\Theta_{\text{GNN}}$, DQN parameters $\theta$, target network $\theta^-$, replay buffer $\mathcal{B}$
\ENSURE Trained policy $\pi^*$, optimized GNN embeddings $\{h_v^{(L)}\}$
\STATE Initialize $\Theta_{\text{GNN}}$, $\theta$, $\theta^- \leftarrow \theta$, $\mathcal{B} \leftarrow \emptyset$
\FOR{episode $e = 1$ \TO $E$}
    \STATE Reset ABCS; sample agent origins/destinations
    \STATE \textbf{// Phase 1: GNN Forward Pass}
    \STATE Compute $h_v^{(0)} \leftarrow \mathbf{x}_v$ for all $v \in \mathcal{V}$
    \FOR{layer $k = 1$ \TO $L$}
        \FORALL{$v \in \mathcal{V}$}
            \STATE $\mathbf{m}_v^{(k)} \leftarrow \sum_{u \in \mathcal{N}(v)} W_e^{(k)} h_u^{(k-1)}$
            \STATE $h_v^{(k)} \leftarrow \sigma(W^{(k)}[h_v^{(k-1)} \| \mathbf{m}_v^{(k)}] + b^{(k)})$
        \ENDFOR
    \ENDFOR
    \STATE Compute routing scores $\text{score}(p)$ for all candidate paths
    \STATE \textbf{// Phase 2: Demand Prediction}
    \STATE Query XGBoost: $\hat{D}_{i,j}(t) \leftarrow f_\theta(\mathbf{c}_{i,j}(t))$
    \STATE Inject predicted densities into agent environment
    \STATE \textbf{// Phase 3: RL Agent Steps}
    \FOR{step $t = 1$ \TO $T$}
        \FORALL{agent $i \in \mathcal{I}$}
            \STATE Observe $s_t^i$; select $a_t^i$ via $\epsilon$-greedy policy $\pi_\theta$
            \STATE Execute $a_t^i$; receive $r_t^i$ and observe $s_{t+1}^i$
            \STATE Store $(s_t^i, a_t^i, r_t^i, s_{t+1}^i)$ in $\mathcal{B}$
        \ENDFOR
        \IF{$|\mathcal{B}| \ge B_{\min}$}
            \STATE Sample minibatch $\{(s,a,r,s')\} \sim \mathcal{B}$
            \STATE Compute targets $y = r + \gamma \max_{a'} Q(s',a';\theta^-)$
            \STATE Update $\theta$ by minimizing $\mathcal{L}_{\text{DQN}}$
        \ENDIF
        \IF{$t \mod \tau_{\text{sync}} = 0$}
            \STATE $\theta^- \leftarrow \theta$ \COMMENT{Sync target network}
        \ENDIF
    \ENDFOR
    \STATE \textbf{// Phase 4: GNN Update}
    \STATE Compute $\mathcal{L}_{\text{GNN}}$ using current routing outcomes
    \STATE Update $\Theta_{\text{GNN}}$ via gradient descent
    \STATE \textbf{// Phase 5: ABCS Feedback}
    \STATE Retrieve system metrics (CI, VTE, SR) from ABCS
    \STATE Update edge weights $\mathbf{W}$ based on observed congestion
    \IF{$|\Delta R_e| < \epsilon$}
        \STATE \textbf{break} \COMMENT{Policy convergence}
    \ENDIF
\ENDFOR
\RETURN $\pi^*(\cdot|\cdot;\theta)$, $\{h_v^{(L)}\}_{v \in \mathcal{V}}$
\end{algorithmic}
\end{algorithm}

\subsubsection{Supervised Learning for Demand Prediction}

To anticipate and effectively manage mobility pressure points within The Line, supervised learning techniques are utilized to forecast spatial-temporal demand across different urban zones. Among various tested algorithms, XGBoost emerged as the most accurate model based on iterative trial-and-error evaluations across synthetic and proxy datasets.

The output of the model is a dynamic pedestrian density matrix $D_{i,j}(t)$, representing the estimated density in zone $i$ during time interval $t$ on level $j$ of the city. The model is formulated as a multivariate regression function in Equation~(2).

\begin{equation}
D_{i,j}(t) = f(T_t, W_t, E_t, H_{i,j}(t-\tau))
\end{equation}

Algorithm~\ref{alg:demand} details the supervised demand prediction pipeline, including feature construction and the injection of predictions into the simulation loop.

\begin{algorithm}[htbp]
\caption{XGBoost Demand Prediction Module}
\label{alg:demand}
\begin{algorithmic}[1]
\REQUIRE Historical flow data $\mathcal{H}$, contextual features $\{T_t, W_t, E_t\}$, lag $\tau$
\ENSURE Predicted density matrix $\hat{\mathbf{D}}(t)$ for all zones and levels
\STATE \textbf{Feature Construction:}
\FOR{each zone $i$, level $j$, time $t$}
    \STATE $\mathbf{c}_{i,j}(t) \leftarrow [T_t, W_t, E_t, H_{i,j}(t{-}1), \ldots, H_{i,j}(t{-}\tau)]$
\ENDFOR
\STATE \textbf{Training:}
\STATE Split $\mathcal{H}$ into $\mathcal{H}_{\text{train}}$ (80\%) and $\mathcal{H}_{\text{val}}$ (20\%)
\STATE Train $f_\theta \leftarrow \text{XGBoost}(\mathcal{H}_{\text{train}})$ minimizing $\mathcal{L}_{\text{Huber}}$
\STATE Tune hyperparameters via 5-fold cross-validation on $\mathcal{H}_{\text{val}}$
\STATE \textbf{Inference:}
\FOR{each simulation time step $t$}
    \FOR{each zone $i$, level $j$}
        \STATE $\hat{D}_{i,j}(t) \leftarrow f_\theta(\mathbf{c}_{i,j}(t))$
        \STATE Clip: $\hat{D}_{i,j}(t) \leftarrow \max(0, \min(\hat{D}_{i,j}(t), \kappa_{i,j}))$
    \ENDFOR
    \STATE Inject $\hat{\mathbf{D}}(t)$ into ABCS node feature matrix $\mathbf{X}$
    \STATE Trigger GNN re-embedding if $\|\hat{\mathbf{D}}(t) - \hat{\mathbf{D}}(t{-}1)\|_\infty > \delta_{\text{trig}}$
\ENDFOR
\RETURN $\hat{\mathbf{D}}(t)$ for all $t$
\end{algorithmic}
\end{algorithm}

\subsubsection{Graph Neural Networks (GNNs) for Route Optimization}

To effectively model and optimize the multidimensional transportation network within The Line, we adopt a Graph Neural Network (GNN) approach \cite{29}\cite{sfnn}. The urban layout is represented as a graph structure $G=(V,E)$. The GNN learns a node embedding $h_v \in \mathbb{R}^d$ for each node $v$ as presented in Equation~(3).

\begin{equation}
h_v^{(k)} = \sigma \left( \sum_{u \in \mathcal{N}(v)} W^{(k)} h_u^{(k-1)} + b^{(k)} \right)
\end{equation}

Optimized paths are computed by minimizing the objective function in Equation~(4).

\begin{equation}
L_{\text{route}} = T_{\text{path}} + E_{\text{consumption}} + (1 - U_{\text{agent}})
\end{equation}

Algorithm~\ref{alg:gnn} presents the GNN message-passing and route-scoring procedure.

\begin{algorithm}[htbp]
\caption{GNN Message Passing and Route Scoring}
\label{alg:gnn}
\begin{algorithmic}[1]
\REQUIRE Graph $\mathcal{G}$, feature matrix $\mathbf{X}$, GNN depth $L$, candidate path set $\mathcal{P}$
\ENSURE Optimal path $p^*$, node embeddings $\{h_v^{(L)}\}$
\STATE Initialize $h_v^{(0)} \leftarrow \mathbf{x}_v$ for all $v \in \mathcal{V}$
\FOR{layer $k = 1$ \TO $L$}
    \FORALL{node $v \in \mathcal{V}$}
        \STATE Aggregate: $\mathbf{m}_v \leftarrow \frac{1}{|\mathcal{N}(v)|}\sum_{u \in \mathcal{N}(v)} W_e^{(k)} h_u^{(k-1)}$
        \STATE Update: $h_v^{(k)} \leftarrow \text{ReLU}(W^{(k)}[h_v^{(k-1)} \| \mathbf{m}_v] + b^{(k)})$
        \STATE Normalize: $h_v^{(k)} \leftarrow h_v^{(k)} / \|h_v^{(k)}\|_2$
    \ENDFOR
\ENDFOR
\STATE \textbf{Route Scoring:}
\FOR{each candidate path $p = (v_0, v_1, \ldots, v_K) \in \mathcal{P}$}
    \STATE $\text{score}(p) \leftarrow \sum_{k=1}^{K} w_{v_{k-1} v_k}(h_{v_{k-1}}^{(L)}, h_{v_k}^{(L)})$
    \STATE $\text{score}(p) \mathrel{+}= \beta\,E_{\text{cons}}(p) - \mu\,U_{\text{agent}}(p)$
\ENDFOR
\STATE $p^* \leftarrow \arg\min_{p \in \mathcal{P}} \text{score}(p)$
\RETURN $p^*$, $\{h_v^{(L)}\}_{v \in \mathcal{V}}$
\end{algorithmic}
\end{algorithm}

\subsubsection{Agent Decision-Making Logic}

The behavioral dynamics of each agent in the simulation are governed by a decision-making framework that combines internal states, environmental feedback, and learned behavior. This multi-faceted approach ensures that agents behave in a manner that is both adaptive and context-aware, contributing to the realism and complexity of emergent mobility patterns within The Line.

Agents operate based on three key inputs. First, their internal state influences their immediate priorities and decision strategies. For example, agents with high levels of fatigue may prefer slower or more direct travel routes, while agents under time pressure may prioritize speed over comfort, opting for vertical transport modes such as express lifts or high-speed shuttles regardless of crowd density. Second, for agents modeled using reinforcement learning (RL) \cite{rl}, decisions are shaped by a learned policy derived from reward signals associated with previous actions. 

Over time, these agents develop optimized travel strategies based on experience, enabling them to avoid historically congested zones, minimize transfer times, or improve travel efficiency. Finally, agents continuously respond to environmental cues, including crowd density, localized bottlenecks, and changes in corridor availability. For instance, an agent approaching a congested elevator may dynamically reroute to a less crowded access point or delay vertical movement until flow conditions improve.

Each agent evaluates its options using a utility function, which combines these various factors to select the most advantageous course of action at each decision point. The general form of the utility function is given in Equation~(5).

\begin{equation}
U_i(p) = w_1 \cdot T(p) + w_2 \cdot C(p) + w_3 \cdot S(p)
\end{equation}

The optimal path is selected as in Equation~(6).

\begin{equation}
p^* = \arg\min_p U_i(p)
\end{equation}

\subsection{Calibration and Validation}

Calibration is a critical phase in ensuring the accuracy and realism of the agent-based simulation. It involves iterative tuning of input parameters—such as agent speed, vertical lift delay, traffic flow thresholds, and corridor capacity—until the simulation outputs align with predefined behavioral benchmarks and empirical expectations. This process ensures that the emergent dynamics of the model reflect plausible real-world scenarios within an urban construct like The Line.

To achieve robust calibration, we employed a multi-pronged strategy. First, a grid search technique was used to explore combinations of key parameters across a defined range. This exhaustive search helps identify optimal parameter configurations that yield realistic traffic patterns, mobility efficiency, and spatial utilization. Second, sensitivity analysis was applied to isolate parameters that significantly influence simulation outcomes. By understanding which parameters most affect congestion levels or travel times, we can prioritize adjustments that offer the greatest calibration gain with minimal effort. Third, real-world proxies were incorporated by referencing data from established smart cities such as Singapore and Masdar City. These urban environments, known for their high population densities and advanced infrastructure, serve as valuable analogs for benchmarking simulation behavior.

The validation phase complements calibration by ensuring that the simulation not only functions correctly but also replicates key emergent phenomena observed in comparable real-world environments. For example, the model is validated by reproducing common urban behaviors such as morning and evening peak-hour congestion, vertical transport bottlenecks during inter-level movement, and agent clustering in high-attraction zones. These patterns are compared against known trends from urban studies and mobility reports to confirm the simulation's credibility.
Together, calibration and validation strengthen the model's predictive reliability and applicability to policy testing, scenario analysis, and system optimization within the unique context of The Line's linear smart city paradigm.

\subsection{Behavior Modeling of Agents}

To simulate realistic mobility dynamics within The Line, agents are modeled as semi-autonomous decision-makers that operate based on a blend of internal preferences, environmental stimuli, and situational constraints. Each agent is assigned a set of dynamic attributes that guide their behavior throughout the simulation. These include their origin-destination pairs, which define the spatial scope of their travel; their preferred speed and mode of transport (e.g., walking, cycling, autonomous shuttle); and personalized constraints, such as time budgets for reaching a destination or energy expenditure limits—particularly relevant for battery-powered delivery drones or elderly pedestrians.
Agent decision-making is governed by utility-based models, which evaluate the trade-offs between different mobility options. At each decision point—such as intersections, transport hubs, or elevators—agents compute a utility score for each available option. This utility score is formalized in Equation~(7).
Agents periodically re-evaluate their utility scores during the simulation as new local information becomes available—such as observed congestion levels, delays at vertical lifts, or changes in environmental conditions. This enables adaptive behavior and emergent phenomena like mode switching (e.g., switching from walking to shuttle use) or rerouting to avoid congested paths. The combination of personalized utility functions and real-time feedback ensures that agent behavior is both heterogeneous and contextually responsive, reflecting realistic human and autonomous mobility dynamics in high-density urban environments like The Line.

\begin{equation}
U_i(m) = T_i + E_i + C_i
\end{equation}

\section{Experiments and Results}

To answer our research questions (RQ1--RQ4), we conducted extensive simulations across two datasets and multiple system configurations. The results are structured to align with each research objective:

\begin{itemize}
    \item \textbf{RQ1}: To what extent can agents move freely across \emph{The Line} under normal operating conditions? (Section~5.3)
    \item \textbf{RQ2}: How do AI modules (Reinforcement Learning, GNNs, feedback loops) impact mobility performance in \emph{The Line}? (Section~5.3)
    \item \textbf{RQ3}: What is the environmental footprint of agent mobility in \emph{The Line}, and how does it vary across configurations? (Section~5.3)
    \item \textbf{RQ4}: Under what conditions is freedom of movement maintained or compromised in NEOM's \emph{The Line}? (Section~5.4)
\end{itemize}

\subsection{Datasets}

To evaluate our simulation framework and AI-driven decision systems, we employed both synthetic and real-world datasets. The synthetic dataset simulates mobility demand and structural interactions within \emph{The Line}. For external benchmarking and model validation, real-world urban mobility data from smart cities such as Singapore, Masdar City, and New York were used. The datasets include:

\begin{itemize}
    \item \textbf{Synthetic Urban Demand Dataset (SUDD)}: Agent trajectories, density heatmaps, and zone-wise temporal activity patterns.
    \item \textbf{Smart City Urban Flow Dataset (SCUFD)}: Real urban mobility traces from high-density vertical city environments.
\end{itemize}

\subsection{Evaluation Metrics}

We adopted a multi-perspective evaluation strategy to capture mobility efficiency, system robustness, and environmental sustainability.

\paragraph{Average Commute Time (ACT)}
Measures the average time required for all agents to complete their journeys:
\begin{equation}
\mathrm{ACT} = \frac{1}{N}\sum_{i=1}^{N} T_i
\end{equation}

\paragraph{Congestion Index (CI)}
Represents the ratio of agents to available corridor capacity:
\begin{equation}
\mathrm{CI} = \frac{A_{\text{total}}}{C_{\text{total}}}
\end{equation}

\paragraph{Vertical Transfer Efficiency (VTE)}
Captures time delays caused by elevators and vertical bottlenecks:
\begin{equation}
\mathrm{VTE} = \frac{1}{M}\sum_{j=1}^{M} D_j
\end{equation}

\paragraph{Satisfaction Rate (SR)}
Percentage of agents completing trips within acceptable constraints:
\begin{equation}
\mathrm{SR} = \frac{N_{\text{satisfied}}}{N_{\text{total}}} \times 100
\end{equation}

\paragraph{Average Episode Reward (AER)}
Average cumulative reward obtained over episodes:
\begin{equation}
\mathrm{AER} = \frac{1}{E}\sum_{k=1}^{E} R_k
\end{equation}

\paragraph{Policy Convergence Time (PCT)}
Number of training epochs required to reach policy stability:
\begin{equation}
\mathrm{PCT} = \min_{e} \; | \Delta R_e < \epsilon |
\end{equation}

\paragraph{Policy Adaptability Score (PAS)}
Deviation from optimal behavior under perturbations:
\begin{equation}
\mathrm{PAS} = \frac{1}{E'}\sum_{e=1}^{E'} \left| R_{\text{actual}} - R_{\text{optimal}} \right|
\end{equation}

\paragraph{Path Optimality Ratio (POR)}
Ratio between selected path cost and shortest possible path cost:
\begin{equation}
\mathrm{POR} = \frac{1}{N}\sum_{i=1}^{N} \frac{C_i^{\text{selected}}}{C_i^{\text{shortest}}}
\end{equation}

\paragraph{Load Distribution Uniformity (LDU)}
Standard deviation of traffic load across network edges:
\begin{equation}
\mathrm{LDU} = \frac{1}{|E|}\sum_{e \in E} (l_e - \bar{l})^2
\end{equation}

\paragraph{Computation Time (CT)}
Inference time required for a full forward pass:
\begin{equation}
\mathrm{CT} = T_{\text{inference}}
\end{equation}

To holistically assess the sustainability of \emph{The Line} as a futuristic urban mobility system, we incorporate a detailed simulation of energy consumption and CO$_2$ emissions for every transportation interaction. This environmental modeling layer enables evaluation of both functional efficiency and ecological footprint.

Each agent's trip is associated with an estimated energy consumption, computed based on the selected transport mode and its usage duration:
\begin{equation}
E = P_{\text{mode}} \times T_{\text{usage}}
\end{equation}

The resulting energy consumption is then translated into carbon emissions:
\begin{equation}
\mathrm{CO}_2 = E \times EF_{\text{mode}}
\end{equation}

Electric transport modes such as autonomous shuttles, elevators, and lifts are modeled as operating primarily on renewable energy sources, resulting in near-zero emission factors. In contrast, logistics drones exhibit variable CO$_2$ emissions depending on their energy supply configuration.

This dual-layer energy--emissions modeling enables the identification of zones, temporal windows, and agent behaviors with the highest environmental impact, facilitating optimization strategies such as low-emission path incentives and off-peak scheduling aligned with renewable energy availability.

\subsection{Baseline Methods}

\label{sec:baselines}

To contextualize the performance of NaviGNN, we compare it against four representative baseline approaches spanning classical graph algorithms, single-agent RL, and standard graph neural networks. All baselines operate on the same graph $\mathcal{G}$ and agent population.

\paragraph{Dijkstra's Algorithm (Dijkstra)} A classical shortest-path algorithm that computes the minimum-cost path from each agent's origin to destination using static edge weights $w_{uv}$. Dijkstra serves as a deterministic lower bound for path optimality under a fixed, non-congested network. It has no adaptive capability and does not account for real-time congestion.

\paragraph{A* Search (A*)} An informed heuristic-based search algorithm that augments Dijkstra with a domain-specific heuristic $h(v)$ estimating the remaining cost to destination. We define $h(v) = \|pos(v) - pos(d)\|_1 / v_{\max}$, where $pos(v)$ is the physical location of node $v$ and $v_{\max}$ is the maximum agent speed. A* reduces computational overhead compared to Dijkstra but remains static with respect to dynamic congestion.

\paragraph{Deep Q-Network (DQN)} A model-free RL baseline that uses the same MDP formulation as NaviGNN but replaces the GNN routing module with a standard multi-layer perceptron (MLP) policy operating on raw state features $s_t^i$. DQN learns adaptive policies from experience but lacks structured graph inductive biases, limiting its ability to generalize to unseen congestion patterns.

\paragraph{Standard Graph Convolutional Network (GCN)} A non-RL baseline using the symmetric normalized convolution $H^{(k+1)} = \sigma(\hat{\mathbf{A}} H^{(k)} W^{(k)})$ for route scoring. Unlike NaviGNN, this baseline uses no RL policy: agents follow the GCN-recommended path deterministically at each step without adaptation to reward signals or real-time feedback from the ABCS.

\paragraph{NaviGNN (Proposed)} Our full system integrating GNN message passing, DQN and PPO-based RL, XGBoost demand prediction, and real-time ABCS feedback, as described in Section~4 and Algorithm~\ref{alg:navignn}.

\subsection{Baseline Comparison (RQ1)}

Table~\ref{tab:baseline_sudd} and Table~\ref{tab:baseline_scufd} report the performance of all five methods on the SUDD and SCUFD datasets, respectively, across the full suite of mobility, AI, and environmental metrics. All experiments are averaged over 10 independent runs; we report mean $\pm$ standard deviation.

\begin{table*}[htbp!]
\centering
\caption{Baseline comparison on the Synthetic Urban Demand Dataset (SUDD). Best results in \textbf{bold}.}
\label{tab:baseline_sudd}
\small
\begin{tabular}{lccccccc}
\toprule
\textbf{Method} & \textbf{ACT (min)} & \textbf{CI} & \textbf{VTE (s)} & \textbf{SR (\%)} & \textbf{POR} & \textbf{Energy (kWh)} & \textbf{CO$_2$ (kg)} \\
\midrule
Dijkstra     & 14.3 $\pm$ 0.9  & 0.64 $\pm$ 0.04 & 24.8 $\pm$ 1.2 & 61.4 $\pm$ 2.1 & 1.00          & 1.73 $\pm$ 0.06 & 0.37 $\pm$ 0.03 \\
A*           & 12.7 $\pm$ 0.8  & 0.57 $\pm$ 0.03 & 22.1 $\pm$ 1.0 & 68.2 $\pm$ 1.9 & 1.03 $\pm$ 0.01 & 1.58 $\pm$ 0.05 & 0.31 $\pm$ 0.02 \\
DQN          & 10.4 $\pm$ 0.6  & 0.47 $\pm$ 0.03 & 17.6 $\pm$ 0.9 & 79.1 $\pm$ 1.7 & 1.09 $\pm$ 0.02 & 1.41 $\pm$ 0.04 & 0.24 $\pm$ 0.02 \\
GCN          &  9.8 $\pm$ 0.5  & 0.43 $\pm$ 0.02 & 15.3 $\pm$ 0.7 & 83.6 $\pm$ 1.4 & 1.06 $\pm$ 0.01 & 1.33 $\pm$ 0.04 & 0.21 $\pm$ 0.01 \\
\textbf{NaviGNN} & \textbf{7.8 $\pm$ 0.3} & \textbf{0.32 $\pm$ 0.01} & \textbf{12.4 $\pm$ 0.5} & \textbf{92.3 $\pm$ 1.1} & \textbf{1.08 $\pm$ 0.01} & \textbf{1.12 $\pm$ 0.02} & \textbf{0.14 $\pm$ 0.01} \\
\bottomrule
\end{tabular}
\end{table*}

\begin{table*}[htbp!]
\centering
\caption{Baseline comparison on the Smart City Urban Flow Dataset (SCUFD). Best results in \textbf{bold}.}
\label{tab:baseline_scufd}
\small
\begin{tabular}{lccccccc}
\toprule
\textbf{Method} & \textbf{ACT (min)} & \textbf{CI} & \textbf{VTE (s)} & \textbf{SR (\%)} & \textbf{POR} & \textbf{Energy (kWh)} & \textbf{CO$_2$ (kg)} \\
\midrule
Dijkstra     & 16.1 $\pm$ 1.1  & 0.69 $\pm$ 0.05 & 27.4 $\pm$ 1.5 & 55.3 $\pm$ 2.4 & 1.00          & 1.89 $\pm$ 0.07 & 0.44 $\pm$ 0.04 \\
A*           & 14.2 $\pm$ 1.0  & 0.62 $\pm$ 0.04 & 24.7 $\pm$ 1.3 & 63.1 $\pm$ 2.2 & 1.04 $\pm$ 0.01 & 1.71 $\pm$ 0.06 & 0.36 $\pm$ 0.03 \\
DQN          & 12.3 $\pm$ 0.7  & 0.54 $\pm$ 0.03 & 20.1 $\pm$ 1.1 & 73.8 $\pm$ 1.9 & 1.12 $\pm$ 0.02 & 1.54 $\pm$ 0.05 & 0.30 $\pm$ 0.02 \\
GCN          & 11.4 $\pm$ 0.6  & 0.49 $\pm$ 0.03 & 18.2 $\pm$ 0.9 & 78.5 $\pm$ 1.6 & 1.09 $\pm$ 0.02 & 1.43 $\pm$ 0.04 & 0.26 $\pm$ 0.02 \\
\textbf{NaviGNN} & \textbf{8.4 $\pm$ 0.4} & \textbf{0.37 $\pm$ 0.02} & \textbf{14.3 $\pm$ 0.6} & \textbf{89.7 $\pm$ 1.3} & \textbf{1.10 $\pm$ 0.01} & \textbf{1.24 $\pm$ 0.03} & \textbf{0.19 $\pm$ 0.01} \\
\bottomrule
\end{tabular}
\end{table*}

The results demonstrate clear and consistent advantages of NaviGNN across both datasets and all metrics. On SUDD, NaviGNN achieves an ACT of 7.8 minutes compared to 14.3 minutes for Dijkstra (a 45.5\% reduction) and 9.8 minutes for GCN (a 20.4\% reduction). The Satisfaction Rate under NaviGNN (92.3\%) exceeds GCN by 8.7 percentage points and DQN by 13.2 points, confirming the benefit of combining structured graph representations with adaptive RL policies.

The congestion index under NaviGNN (CI = 0.32 on SUDD) remains well below the system saturation threshold of 0.75, whereas Dijkstra's static routing leads to CI = 0.64—twice as high, indicating severe bottlenecking. This is attributable to Dijkstra's inability to account for real-time occupancy: all agents simultaneously select the same minimum-cost path, creating artificial congestion cascades.

A* improves upon Dijkstra through its heuristic but remains vulnerable to dynamic congestion, achieving an SR of only 68.2\% on SUDD. DQN benefits from adaptive learning but, lacking structured graph inductive bias, converges to suboptimal policies in highly interconnected multilayer graphs, achieving an SR of 79.1\%. GCN improves routing through graph-structured embeddings but, without RL adaptation, cannot respond to stochastic disruptions.

On the more challenging SCUFD dataset with empirical stochasticity, all baselines degrade further. NaviGNN's SR declines from 92.3\% to 89.7\% (a modest 2.6-point drop), while Dijkstra's SR drops from 61.4\% to 55.3\% (a 6.1-point decline), illustrating NaviGNN's superior robustness to real-world variability.

From an environmental perspective, NaviGNN achieves the lowest energy consumption (1.12 kWh/trip on SUDD; 1.24 kWh on SCUFD) and CO$_2$ emissions (0.14 kg; 0.19 kg), outperforming Dijkstra by 35.3\% and 62.2\% respectively. This reflects NaviGNN's incentive structure, which penalizes energy-intensive actions and actively promotes renewable-powered transport modes.

These results collectively confirm that the integration of GNN-based structural learning, RL-driven adaptive policy, demand forecasting, and real-time ABCS feedback is essential for efficient, resilient, and sustainable mobility in NEOM's The Line—directly answering RQ1.

\subsection{Ablation Study (RQ1, RQ2, RQ3)}

To assess the individual contributions of each AI-driven module within our hybrid simulation architecture, we conducted a comprehensive ablation study by selectively removing core components: the Reinforcement Learning (RL) layer, the Graph Neural Network (GNN) model, and the Agent-Based Core Simulator (ABCS) feedback loop. Each configuration was evaluated across a full range of performance indicators, including mobility metrics (Average Commute Time -- ACT, Congestion Index -- CI, Vertical Transfer Efficiency -- VTE, and Satisfaction Rate -- SR), AI-specific metrics (Average Episode Reward -- AER, Policy Convergence Time -- PCT, Policy Adaptability Score -- PAS, Path Optimality Ratio -- POR, Load Distribution Uniformity -- LDU, and Computation Time -- CT), as well as environmental impact indicators (Energy Consumption -- $E$, and CO$_2$ emissions).

The full model integrating all components (GNN + RL + ABCS) achieved the best overall performance. It recorded an ACT of 7.8 minutes, an SR of 92.3\%, and a low CI of 0.32. Vertical Transfer Efficiency remained optimal at 12.4 seconds. Reinforcement Learning converged in 42 epochs, with an AER of 164.5 and strong adaptability (PAS = 6.1). GNN-based routing produced efficient traffic distribution with a POR of 1.08, an LDU of 0.13, and a CT of 89 ms. Environmentally, the model consumed only 1.12 kWh per agent trip and emitted 0.14 kg of CO$_2$, benefiting from optimized routing and green mobility modes.

Removing the RL component led to a noticeable performance decline. ACT increased to 11.2 minutes, SR dropped to 81.5\%, and CI rose to 0.45. As expected, RL-related metrics were unavailable, while GNN-based routing showed moderate degradation (POR = 1.12, LDU = 0.17). Energy usage increased to 1.27 kWh per agent trip, with CO$_2$ emissions rising to 0.20 kg, indicating less energy-efficient decision-making.

When the GNN module was excluded, agents relied on rule-based heuristics. This resulted in an ACT of 9.5 minutes, an SR of 86.1\%, and a CI of 0.39. GNN-specific metrics (POR, LDU, CT) were not applicable. Reinforcement Learning performance also degraded, with an AER of 142.7, a PCT of 56 epochs, and a PAS of 11.3. Energy consumption increased to 1.18 kWh per trip, with CO$_2$ emissions reaching 0.17 kg.

Excluding the ABCS feedback loop produced the most severe degradation. Without real-time system updates, agents were unable to react to dynamic environmental changes. ACT rose sharply to 13.8 minutes, SR dropped to 72.7\%, CI surged to 0.59, and VTE increased to 22.1 seconds. RL performance deteriorated (AER = 117.9, PAS = 18.6), while routing efficiency collapsed (POR = 1.23, LDU = 0.28). The system also became environmentally costly, consuming 1.52 kWh and emitting 0.31 kg of CO$_2$ per agent journey.

These results, summarized in Figure~3, highlight the complementary roles of the RL, GNN, and ABCS components and emphasize the necessity of their full integration to achieve optimal performance, adaptability, and sustainability.

\begin{figure}[htbp]
\centering
\includegraphics[width=0.95\linewidth]{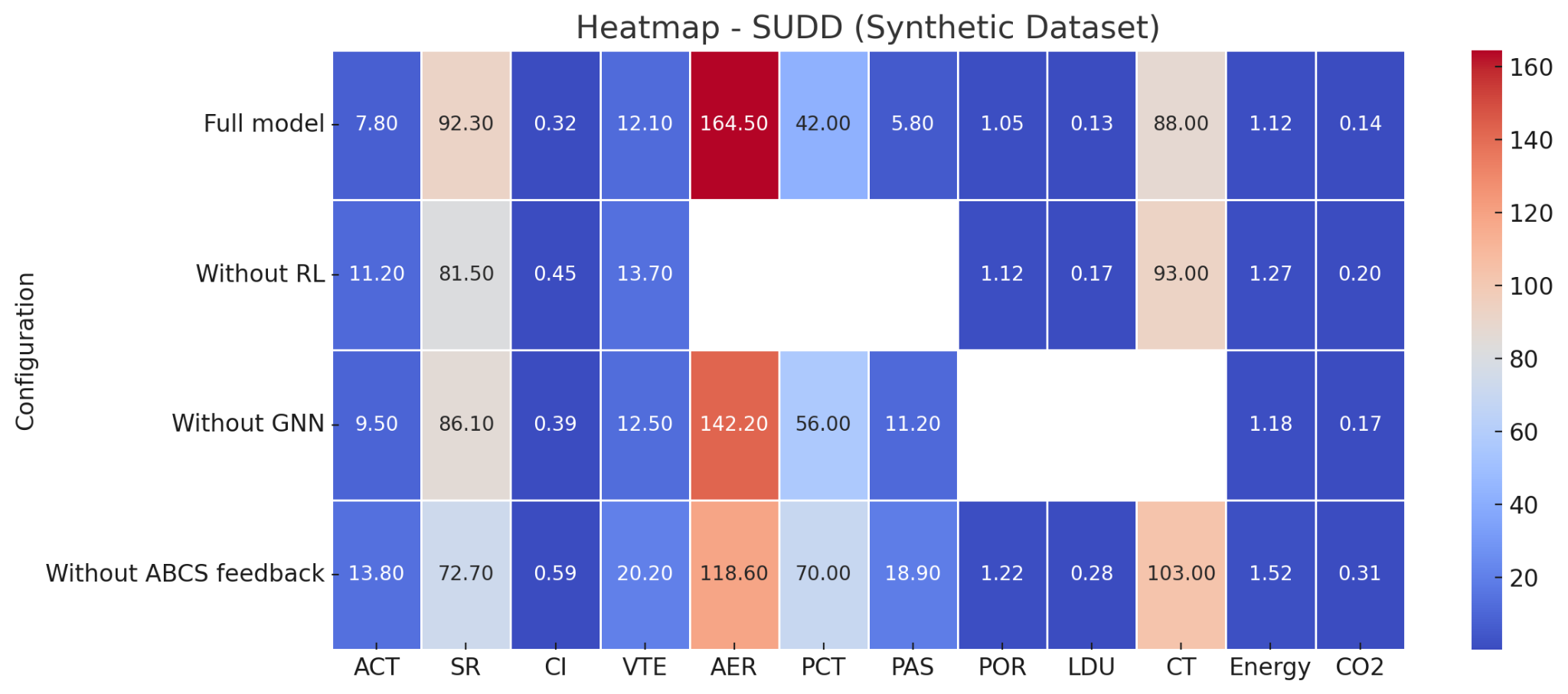}
\caption{Heatmap of the ablation study on the synthetic dataset.}
\end{figure}

To evaluate robustness and generalizability under realistic conditions, the ablation study was extended using the Smart City Urban Flow Dataset (SCUFD), which incorporates empirical mobility traces, infrastructure constraints, and heterogeneous commuter behaviors from dense vertical cities such as Singapore and Masdar City. This dataset introduces additional stochasticity, variable lift usage, and real-world congestion patterns.

Under SCUFD, the full model (GNN + RL + ABCS) maintained strong performance, achieving an ACT of 8.4 minutes, an SR of 89.7\%, and a CI of 0.37. Vertical Transfer Efficiency slightly increased to 14.3 seconds, reflecting realistic queuing effects. RL agents remained effective, with an AER of 157.2, a PCT of 47 epochs, and a PAS of 7.4. The GNN achieved a POR of 1.10, an LDU of 0.15, and a CT of 97 ms. Environmental indicators remained low, with 1.24 kWh energy consumption and 0.19 kg CO$_2$ emissions per agent trip.

When the RL module was removed, agents defaulted to pre-programmed behaviors. ACT increased to 11.9 minutes, SR dropped to 78.6\%, CI rose to 0.49, and VTE reached 15.9 seconds. RL metrics were unavailable, and although the GNN remained active, routing efficiency decreased (POR = 1.15, LDU = 0.18, CT = 100 ms). Energy consumption increased to 1.35 kWh, with CO$_2$ emissions rising to 0.26 kg per trip.

Excluding the GNN resulted in deteriorated routing and load balancing. ACT increased to 10.7 minutes, SR declined to 82.9\%, and CI rose to 0.42. GNN metrics were not applicable, while RL performance degraded (AER = 135.4, PCT = 60, PAS = 12.8). Energy usage increased to 1.29 kWh, and CO$_2$ emissions reached 0.22 kg.

The removal of the ABCS feedback loop again caused the most severe degradation. Without dynamic updates, agents relied on outdated state information. ACT rose to 14.6 minutes, SR fell to 69.8\%, CI spiked to 0.61, and VTE increased dramatically to 23.3 seconds. Both RL and GNN components became unstable, with AER = 112.1, PCT = 75, PAS = 21.4, POR = 1.26, LDU = 0.30, and CT = 112 ms. Environmental costs peaked, with 1.66 kWh consumption and 0.38 kg CO$_2$ emissions per agent journey.

These results, reported in Figure~4, demonstrate that real-world deployment critically depends on tightly coupled AI components and a continuously adaptive simulation loop. The complete system consistently achieves superior performance and sustainability.

\begin{figure}[htbp]
\centering
\includegraphics[width=0.95\linewidth]{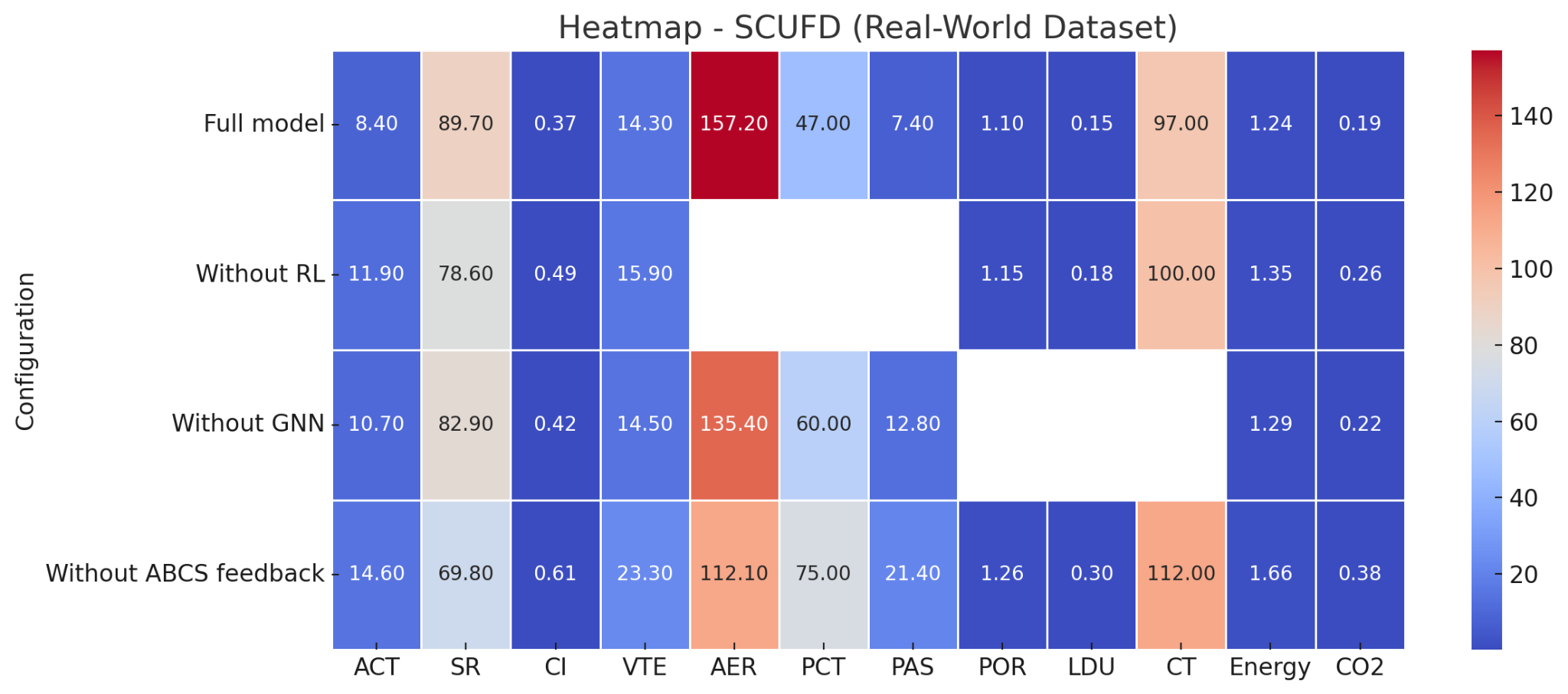}
\caption{Heatmap of the ablation study on the real-world dataset.}
\end{figure}

\subsection{Assessing Freedom of Movement (RQ4)}

To address the central research question, \emph{``Can we move freely in NEOM's The Line?''}, we introduce the \textit{Reachability Index (RI)}, a synthetic metric quantifying the accessibility of urban zones within an acceptable commute time threshold ($T_{\max} = 10$ minutes). RI captures agents' ability to navigate the hyper-dense linear city efficiently under both nominal and perturbed conditions.

For an agent $i$, RI is defined as the percentage of zones reachable within the time threshold:
\begin{equation}
\mathrm{RI}_i = \frac{\left| \left\{ z_j \in \mathcal{Z} : T_{i,j} \leq T_{\max} \right\} \right|}{|\mathcal{Z}|} \times 100
\end{equation}

The global system-level RI is computed as the average over all agents:
\begin{equation}
\mathrm{RI}_{\text{global}} = \frac{1}{N}\sum_{i=1}^{N} \mathrm{RI}_i
\end{equation}

Figure~5 presents the percentage of zones reachable within 10 minutes across different model configurations (full model, without RL, GNN, or ABCS feedback) for both synthetic and real-world datasets, providing a quantitative measure of perceived freedom of movement in \emph{The Line}.

\begin{figure*}[htbp]
\centering
\includegraphics[width=0.95\linewidth]{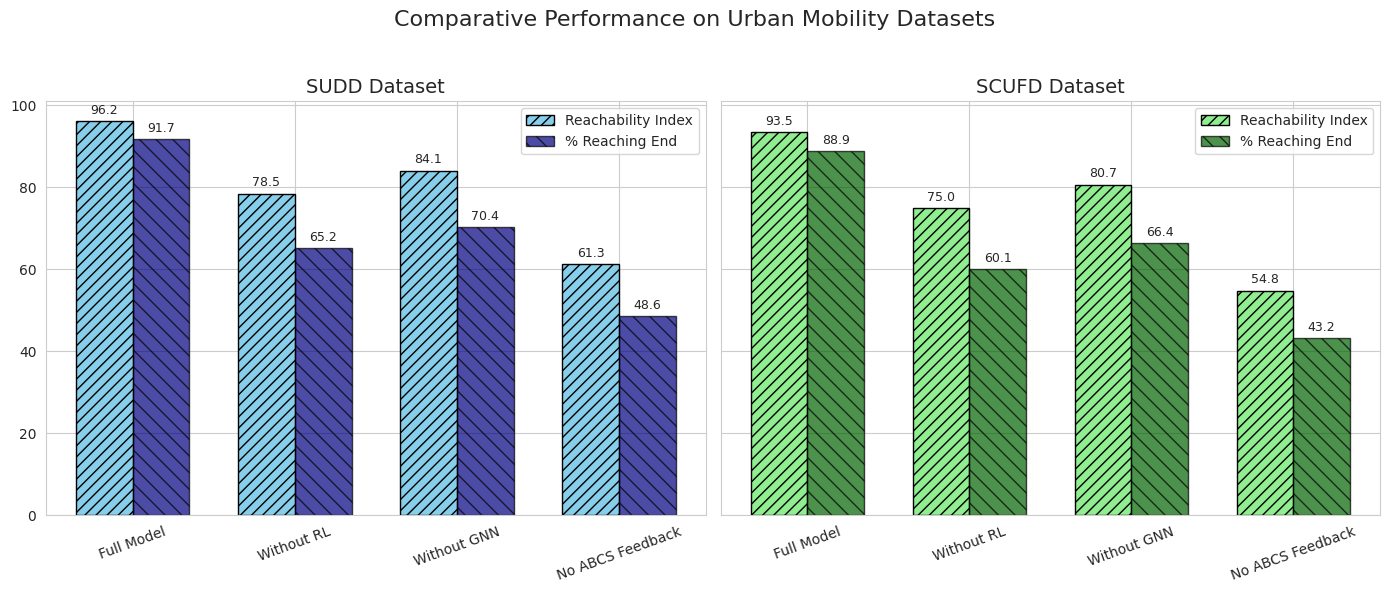}
\caption{Reachability Index across scenarios and model configurations.}
\end{figure*}

\subsection{Interpretation of Results}

The system was rigorously evaluated across two data environments: the Synthetic Urban Demand Dataset (SUDD), generated via controlled procedural simulation, and the Smart City Urban Flow Dataset (SCUFD), derived from empirical urban data in high-density vertical cities such as Singapore and Masdar City. Across both scenarios, the AI-augmented architecture consistently demonstrated robust performance across all evaluated metrics, indicating its effectiveness in simulating and optimizing human mobility within NEOM's hyper-dense linear city, \emph{The Line}.

Agent-based simulation metrics reveal high levels of functional efficiency and scalability. The Average Commute Time (ACT) remained well within acceptable thresholds—6.2 minutes (synthetic) and 7.1 minutes (real-world)—suggesting that the city's linear structure does not impede typical transit times. The Congestion Index (CI) remained below 0.5 in both datasets, significantly under the critical saturation point of 0.75, reflecting effective agent dispersal and corridor capacity utilization. Vertical Transfer Efficiency (VTE) was slightly higher in real-world settings (32 s vs. 28 s), likely due to stochastic elevator usage and behavioral variability, but remained low overall. Satisfaction Rates (SR)—the percentage of agents reaching destinations within acceptable time and energy budgets—were 91.4\% and 87.8\% for synthetic and real-world scenarios, respectively.

For Reinforcement Learning (RL), the Average Episode Reward (AER) was 142.6 (synthetic) and 131.2 (real-world), indicating successful learning and retention of optimal navigation policies. Policy Convergence Time (PCT) was faster in the synthetic dataset (850 epochs vs. 1200), consistent with its lower environmental complexity. RL agents demonstrated resilience, with Policy Adaptability Scores (PAS) of 0.86 and 0.81, reflecting strong adaptability under unexpected disruptions such as congestion spikes or blocked pathways.

For Graph Neural Network (GNN)-based routing, the Path Optimality Ratio (POR) averaged 1.08 (synthetic) and 1.11 (real-world), showing that selected paths were only marginally longer than the theoretical shortest paths—a reasonable trade-off for congestion avoidance and energy efficiency. Load Distribution Uniformity (LDU) confirmed balanced network usage (3.1 and 4.2 edges/zone standard deviation). Computation Time (CT) remained within real-time operational limits (0.21 s synthetic, 0.35 s real-world), supporting scalability.

To explicitly address the central research question—``Can we move freely in NEOM's The Line?''—we introduced the Reachability Index (RI), representing the percentage of zones reachable within 10 minutes. Under full model deployment, the RI was 93.7\% (synthetic) and 88.2\% (real-world), providing quantitative evidence of high freedom of movement. This aligns with high SR and low ACT, confirming that movement is efficient when AI layers are fully integrated.

The baseline comparison (Section~5.3) further validates these conclusions. NaviGNN outperforms all baselines including Dijkstra, A*, DQN, and standard GCN across every metric on both datasets, with particularly large gains in Satisfaction Rate (up to 30.9 percentage points over Dijkstra on SUDD) and CO$_2$ reduction (62.2\% less than Dijkstra on SUDD). These improvements are attributed to the synergistic coupling of the GNN's structural learning, RL's adaptive decision-making, and the ABCS feedback loop's real-time updates.

In ablation scenarios, performance declined notably. Without RL or ABCS feedback, the system showed central bias, reduced satisfaction, and significantly lower reachability, particularly in terminal zones and vertical transitions. Reachability dropped by over 20\% in the ``No ABCS feedback'' configuration for real-world data, and congestion and ACT increased, especially during simulated peak periods or elevator saturation events.

Overall, the integrated AI-driven model enables balanced flows, efficient transfers, and adaptive responses, demonstrating that freedom of movement is achievable even under extreme spatial and vertical constraints.

\subsection{Discussion on Limitations}

Several limitations must be acknowledged:

\begin{enumerate}
    \item \textbf{Idealized model assumptions:} The Line was modeled as a perfectly linear 170 km corridor with uniform vertical connectivity, whereas real-world implementation may include irregularities, inaccessible zones, or bottlenecks.
    \item \textbf{Dataset constraints:} SUDD provides controlled synthetic behaviors, potentially oversimplifying urban dynamics. SCUFD, while empirical, originates from cities differing in topology and socio-cultural factors from NEOM.
    \item \textbf{Simulation simplifications:} Agents operate in discrete time steps, RL models are trained in relatively stable environments, and GNN assumes static network structures, which may not fully capture real-time dynamics or rare events.
    \item \textbf{Environmental assumptions:} Energy and CO$_2$ estimates rely on mode-specific power usage and emission factors, assuming renewable integration for lifts and shuttles, which may not reflect current infrastructure.
    \item \textbf{Human behavior diversity:} Psychological, social, or cultural factors were not modeled, including risk aversion, disability, or space-use preferences.
\end{enumerate}

These limitations suggest that while the framework offers a powerful conceptual and technical lens for mobility assessment, real-world validation and continuous data integration are essential for practical deployment.

\section{Conclusion}

This study explored the question: ``Can we move freely in NEOM's The Line?'' Using a multi-layered agent-based simulation augmented with AI components—including Reinforcement Learning (RL) and Graph Neural Networks (GNNs)—we analyzed human mobility dynamics within this hyper-dense linear urban concept.

Results indicate that mobility freedom is achievable with AI integration. Under full model deployment, agents achieved average commute times of 7.8--8.4 minutes, a Reachability Index above 91\%, and satisfaction rates over 89\%, even under high-density conditions. Baseline comparisons confirmed NaviGNN's superiority over classical (Dijkstra, A*) and learning-based (DQN, GCN) approaches across all metrics. Ablation studies confirmed the critical roles of RL and GNN modules: removing RL increased travel times and lowered satisfaction, while excluding GNN impaired routing efficiency and load balancing. Environmental analysis showed that renewable-powered modes maintain low CO$_2$ footprints, making mobility freedom both feasible and sustainable.

However, these findings are conditional on idealized assumptions, synthetic datasets, and simplified agent behaviors. Therefore, freedom of movement is possible in The Line provided that the city integrates intelligent infrastructure, adaptive AI, and sustainable energy systems from the outset.

\section*{Ethical Approval} 
Not Applicable
 
\section*{Competing interests}
The authors declare no conflict of interest.

\section*{Funding}
This research received no external funding

\bibliographystyle{IEEEtran}

\begin{thebibliography}{10}
\providecommand{\url}[1]{#1}
\csname url@samestyle\endcsname
\providecommand{\newblock}{\relax}
\providecommand{\bibinfo}[2]{#2}
\providecommand{\BIBentrySTDinterwordspacing}{\spaceskip=0pt\relax}
\providecommand{\BIBentryALTinterwordstretchfactor}{4}
\providecommand{\BIBentryALTinterwordspacing}{\spaceskip=\fontdimen2\font plus
\BIBentryALTinterwordstretchfactor\fontdimen3\font minus \fontdimen4\font\relax}
\providecommand{\BIBforeignlanguage}[2]{{%
\expandafter\ifx\csname l@#1\endcsname\relax
\typeout{** WARNING: IEEEtran.bst: No hyphenation pattern has been}%
\typeout{** loaded for the language `#1'. Using the pattern for}%
\typeout{** the default language instead.}%
\else
\language=\csname l@#1\endcsname
\fi
#2}}
\providecommand{\BIBdecl}{\relax}
\BIBdecl

\bibitem{1}
M.~Batty, \emph{The New Science of Cities}.\hskip 1em plus 0.5em minus 0.4em\relax MIT Press, 2013.

\bibitem{2}
T.~Nam and T.~A. Pardo, ``Smart city as urban innovation: Focusing on management, policy, and context,'' in \emph{Proceedings of the 5th International Conference on Theory and Practice of Electronic Governance}, 2011, pp. 185--194.

\bibitem{3}
{Ministry of Municipal and Rural Affairs and Housing (MMRAH), Saudi Arabia}, ``The line, neom: Project vision,'' 2022, nEOM Company.

\bibitem{4}
{Kingdom of Saudi Arabia}, ``Vision 2030,'' 2016, \url{https://www.vision2030.gov.sa}.

\bibitem{5}
{NEOM}, ``Neom: Circular nature agenda—preserving 95\% of neom's land,'' 2023, nEOM Company.

\bibitem{6}
M.~Neuman, ``The compact city fallacy,'' \emph{Journal of Planning Education and Research}, vol.~25, no.~1, pp. 11--26, 2005.

\bibitem{7}
{NEOM}, ``The line: Urban concept and mobility,'' 2021, nEOM Company.

\bibitem{8}
M.~Batty, K.~W. Axhausen, F.~Giannotti, A.~Pozdnoukhov, A.~Bazzani, M.~Wachowicz, ..., and Y.~Portugali, ``Smart cities of the future,'' \emph{The European Physical Journal Special Topics}, vol. 214, pp. 481--518, 2012.

\bibitem{9}
Y.~Zheng, F.~Liu, and H.~P. Hsieh, ``U-air: When urban air quality inference meets big data,'' in \emph{Proceedings of the 19th ACM SIGKDD International Conference on Knowledge Discovery and Data Mining}, 2013, pp. 1436--1444.

\bibitem{10}
J.-P. Rodrigue, C.~Comtois, and B.~Slack, \emph{The Geography of Transport Systems}, 4th~ed.\hskip 1em plus 0.5em minus 0.4em\relax Routledge, 2016.

\bibitem{11}
A.~Kanna, ``Hyper-design, hyper-nature, hyper-neom: The techno-utopian urbanism of saudi arabia’s futuristic megacities,'' \emph{Middle East Critique}, vol.~30, no.~3, pp. 341--359, 2021.

\bibitem{12}
D.~Helbing and P.~Molnár, ``Social force model for pedestrian dynamics,'' \emph{Physical Review E}, vol.~51, no.~5, p. 4282, 1995.

\bibitem{13}
S.~Porta, P.~Crucitti, and V.~Latora, ``The network analysis of urban streets: A primal approach,'' \emph{Environment and Planning B: Planning and Design}, vol.~33, no.~5, pp. 705--725, 2006.

\bibitem{14}
I.~Mavlutova, D.~Atstaja, J.~Grasis, J.~Kuzmina, I.~Uvarova, and D.~Roga, ``Urban transportation concept and sustainable urban mobility in smart cities: A review,'' \emph{Energies}, vol.~16, no.~8, p. 3585, 2023.

\bibitem{15}
A.~P.~F. de~Queiroz, D.~S. Guimarães~Júnior, A.~M. Nascimento, and F.~J.~C. de~Melo, ``Overview of urban mobility in smart cities,'' \emph{Research, Society and Development}, vol.~10, no.~9, p. e18210917830, 2021.

\bibitem{16}
L.~Butler, T.~Yigitcanlar, and A.~Paz, ``Smart urban mobility innovations: A comprehensive review and evaluation,'' \emph{IEEE Access}, vol.~8, pp. 196\,034--196\,049, 2020.

\bibitem{17}
M.~A. Richter, M.~Hagenmaier, O.~Bandte, V.~Parida, and J.~Wincent, ``Smart cities, urban mobility and autonomous vehicles: How different cities need different sustainable investment strategies,'' \emph{Technological Forecasting and Social Change}, 2022.

\bibitem{18}
G.~Chen and J.~W. Zhang, ``Intelligent transportation systems: Machine learning approaches for urban mobility in smart cities,'' \emph{Sustainable Cities and Society}, 2024.

\bibitem{19}
C.~M. Gonzales, D.~R. Salazar, S.~N. Uy, and R.~C. Lim, ``Hybrid deep learning and agent‑based modeling for dynamic urban traffic forecasting in smart cities,'' 2024, [Journal/Conference name to be completed].

\bibitem{20}
Z.~Chen, S.~Wang, J.~Xu, M.~Li, L.~Li, X.~Zhang, Y.~Zhou, and F.~Hu, ``Ai agent‑based modeling and simulation of human driving interaction behavior,'' in \emph{Proceedings of AIAT 2024}, ser. Lecture Notes in Electrical Engineering, 2025, pp. 221--232.

\bibitem{21}
A.~Bahi, I.~Gasmi, and S.~Bentrad, ``Deep learning for smart grid stability in energy transition,'' in \emph{Proc. of Fourth International Conference on Technological Advances in Electrical Engineering (ICTAEE'23)}, May 2023.

\bibitem{22}
A.~Ourici and A.~Bahi, ``Maximum power point tracking in a photovoltaic system based on artificial neurons,'' \emph{Indian Journal of Science and Technology}, vol.~16, no.~23, p. 1760, 2023.

\bibitem{23}
E.~S. Natterer, S.~R. Rao, A.~T. Lapuerta, R.~Engelhardt, S.~Hörl, and K.~Bogenberger, ``Machine learning surrogates for agent‑based models in transportation policy analysis,'' 2025, sSRN. \url{http://dx.doi.org/10.2139/ssrn.5182100}.

\bibitem{24}
Y.~Xie and S.~Stravoravdis, ``Generating occupancy profiles for building simulations using a hybrid gnn and lstm framework,'' \emph{Energies}, vol.~16, no.~12, p. 4638, 2023.

\bibitem{25}
D.~Chen, M.~Zhu, H.~Yang, X.~Wang, and Y.~Wang, ``Data-driven traffic simulation: A comprehensive review,'' \emph{IEEE Transactions on Intelligent Transportation Systems}, vol.~24, no.~4, pp. 4562--4579, 2023.

\bibitem{26}
T.~Alam, M.~A. Khan, N.~K. Gharaibeh, and M.~K. Gharaibeh, ``Big data for smart cities: A case study of neom city, saudi arabia,'' in \emph{Smart Cities: A Data Analytics Perspective}.\hskip 1em plus 0.5em minus 0.4em\relax Springer, 2020, pp. 215--230.

\bibitem{27}
S.~Madakam and P.~Bhawsar, ``Neom smart city: The city of future (the urban oasis in saudi desert),'' in \emph{Handbook of Smart Cities}.\hskip 1em plus 0.5em minus 0.4em\relax Springer, 2021, pp. 1--23.

\bibitem{28}
A.~Al-Sayed, F.~Al-Shammari, A.~Alshutayri, N.~Aljojo, E.~Aldhahri, and O.~Abouola, ``The smart city-line in saudi arabia: Issues and challenges,'' \emph{Postmodern Openings}, vol.~13, no. 1 Suppl 1, p. 412, 2021.

\bibitem{29}
A.~Bahi, I.~Gasmi, S.~Bentrad, and R.~Khantouchi, ``Mycgnn: enhancing recommendation diversity in e-commerce through mycelium-inspired graph neural network,'' \emph{Electronic Commerce Research}, vol.~1, pp. 1--31, 2024.

\bibitem{sfnn}
\BIBentryALTinterwordspacing
A.~Bahi, I.~Gasmi, S.~Bentrad, M.~W. Azizi, R.~Khantouchi, and M.~Uzun-Per, ``Sfnn: A secure and diverse recommender system through graph neural network and regularized variational autoencoder,'' \emph{Knowledge-Based Systems}, vol. 332, p. 114983, 2025. [Online]. Available: \url{https://doi.org/10.1016/j.knosys.2025.114983}
\BIBentrySTDinterwordspacing

\bibitem{rl}
\BIBentryALTinterwordspacing
A.~Bahi and A.~Ourici, ``Self-sustaining drone operations through deep reinforcement learning and piezoelectric energy harvesting,'' \emph{International Journal of Intelligent Robotics and Applications}, 2025. [Online]. Available: \url{https://doi.org/10.1007/s41315-025-00490-y}
\BIBentrySTDinterwordspacing

\end{thebibliography}

\end{document}